\documentclass{article}

\usepackage{lineno,hyperref}
\usepackage{amsmath}
\usepackage{makecell}
\usepackage{adjustbox}
\usepackage{graphicx}
\usepackage{placeins}
\usepackage[shortlabels]{enumitem}
\usepackage{dblfloatfix}
\usepackage{pifont} 
\usepackage{amssymb}
\usepackage{pdfpages}

\usepackage{subcaption,siunitx,booktabs}
\usepackage[utf8]{inputenc}
\usepackage{fourier} 
\usepackage{array}
\usepackage{makecell}
\usepackage{arydshln} 
\usepackage{changes}
\usepackage{multirow}
\usepackage{textcase}
\usepackage[tablename=TABLE]{caption}
\DeclareCaptionTextFormat{up}{\MakeTextUppercase{#1}}
\usepackage{arxiv}

\usepackage[utf8]{inputenc} 
\usepackage[T1]{fontenc}    
\usepackage{hyperref}       
\usepackage{url}            
\usepackage{booktabs}       
\usepackage{amsfonts}       
\usepackage{nicefrac}       
\usepackage{microtype}      
\usepackage{lipsum}
\usepackage{fancyhdr}       
\usepackage{graphicx}       
\graphicspath{{media/}}     

\newcommand{\xmark}{\ding{55}}

\definechangesauthor[name={Emre Akbas}, color=blue]{EA}

\title{Does depth estimation help object detection?
\thanks{This work was supported by the Scientific and Technological Research Council of
        Turkey (T\"UB\.ITAK) through the project title ``Object Detection in Videos with Deep Neural Networks'' with grant number 117E054. The numerical calculations reported in this paper were fully performed at TUBITAK ULAKBIM, High Performance and Grid Computing Center (TRUBA resources).} 
}

\author{
  Bedrettin Cetinkaya, Sinan Kalkan, Emre Akbas \\
  Middle East Techical University \\
  Ankara, Turkey\\
}

\begin{document}

\newcommand{\CStart}[0]{\begin{tabular}[c]{@{}c@{}}}
\newcommand{\CEnd}[0]{\end{tabular}}
\newcommand\RotText[1]{\fontsize{9}{9}\selectfont \rotatebox[origin=c]{90}{\parbox{2.6cm}{\centering#1}}}

\newcolumntype{C}{@{\hspace{1pt}}c@{\hspace{1pt}}}

\maketitle

\begin{abstract}
Ground-truth  depth, when combined with color data, helps improve object detection accuracy over baseline models that only use color. However, estimated depth does not always yield improvements. Many factors affect the performance of object detection when estimated depth is used. In this paper, we comprehensively investigate these factors with detailed experiments, such as using ground-truth vs. estimated depth, effects of different state-of-the-art depth estimation networks,  effects of using different indoor and outdoor RGB-D datasets as training data for depth estimation,  and different architectural choices for integrating depth to the base object detector network. We propose an early concatenation strategy of depth, which yields higher mAP than previous works' while using significantly fewer parameters. 
\end{abstract}


\title{Does depth estimation help object detection?}


\section{INTRODUCTION}
\label{sec:introduction}
Over the past few years, RGB-D cameras which are able to capture the color and depth information simultaneously have become more affordable and available. These cameras have enabled researchers to use both sources of information for various problems such as scene labeling \cite{ren2012rgb}, object recognition \cite{lai2011large}, odometry estimation \cite{kerl2013robust}, and depth and surface normal estimation \cite{eigen2015predicting}. In object detection, it is well-known that the depth information, when combined with the color information, helps improve detection accuracy over baseline models that use only color \cite{cao2017exploiting, gupta2014learning, hou2018object}. What is lesser known is whether depth as estimated from the color image itself would improve accuracy and how to best integrate it into the detection pipeline. This question becomes more important as ground-truth depth information is not always available in widely used object detection datasets, e.g. PASCAL \cite{everingham2008pascal,everingham2010pascal} and MS-COCO \cite{lin2014microsoft}. 

With the recent developments in single-image depth estimation  \cite{godard2017unsupervised, hu2019revisiting,li2018megadepth,miangoleh2021boosting,ranftl2019towards,zheng2018t2net}, the use of estimated depth for object detection has become more prevalent. In this paper, we explore  different aspects of adding estimated depth information into the object detection pipeline. Specifically, we study the following questions: (i) Should depth be used as it is or is it more useful when transformed into various other encodings such as horizontal disparity, height above ground, surface normal, etc.? (ii) How and when should the depth information be integrated into the object detection pipeline? 

\begin{figure*}[h]
  \centerline{\includegraphics[width=\linewidth]{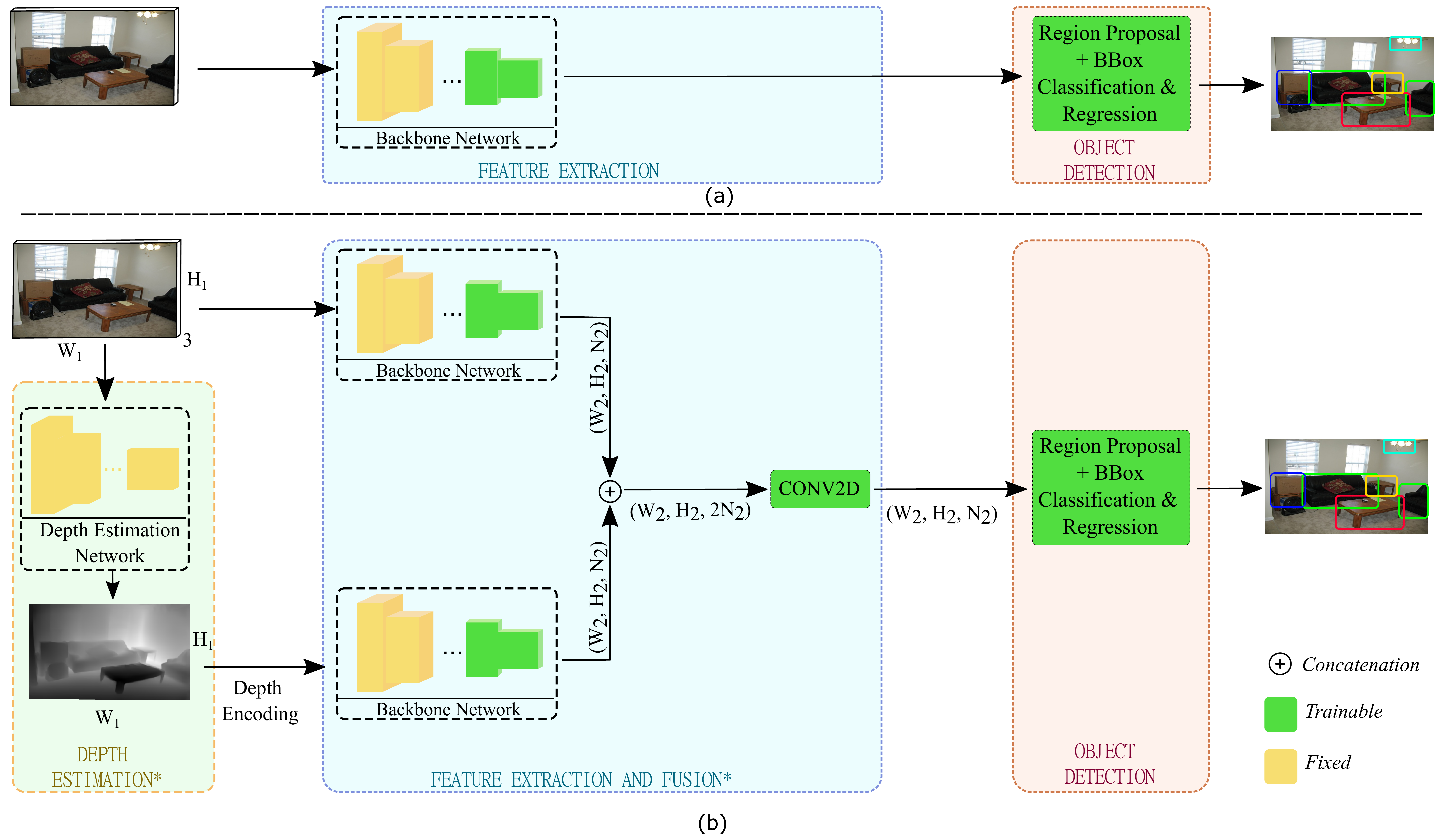}}
  \caption{(a) The main stages in Faster R-CNN. (b) An overview of our modifications in Faster R-CNN. Components marked with an asterisk include our modifications and contributions.  \label{fig:Architecture}}
\end{figure*}

Our work is not the first to integrate estimated depth into object detection. However, to the best of our knowledge, ours presents the most comprehensive exploration of the problem. Previous work either studied the problem  in a limited way \cite{cao2017exploiting}, i.e. they integrated depth only at a later stage, or they increased the number of base object detector parameters at least four times \cite{hou2018object}. In contrast, our model (Figure \ref{fig:Architecture}) which we obtained after thorough experimentation, combines RGB and depth at an early stage (i.e. right after the backbone network) which yields better detection accuracy with a smaller number of parameters compared to previous work. 


In this paper, our contributions are as follows:

\begin{itemize}
\item We provide thorough experimentation on whether/how to process and integrate depth into the object detection pipeline. From our experiments, we conclude that raw estimated depth  should be converted to a depth image, and then they should be passed through a backbone feature extraction network similar to the RGB-image processing branch. An early concatenation of depth and RGB features, i.e. concatenation just after the backbone networks, gives better results than other concatenation points in general. 

\item We experiment with several state-of-the-art single image depth estimation methods and investigate  which  estimator helps object detection the most. 

\item We propose a simple way of integrating depth into a two-stage object detection pipeline, which outperforms the previous methods, while using fewer number of parameters. 
\end{itemize}

The rest of the paper is organized as follows. Section \ref{sect:relatedWork} provides related work on the problem of RGB-D object detection. Section \ref{sect:overview} explains the details of our method and Section \ref{sect:experiments} presents our experiments and their results. Section \ref{sect:conclusion} concludes the paper by providing a discussion on results.


\section{RELATED WORK}
\label{sect:relatedWork}


Object detection is one of the widely studied problems in Computer Vision and a central problem that needs to be addressed in a wide spectrum of applications. 
Deep learning based object detection methods 
(e.g. Region-CNN (R-CNN) \cite{RCNN}, Fast R-CNN \cite{FastRCNN}, Faster R-CNN \cite{FasterRCNN}, SSD \cite{SSD}, YOLO \cite{YOLO}, RetinaNet \cite{RETINANET}), 
can be broadly analyzed in two main categories: (i) Two-stage detectors (e.g., 
Faster R-CNN), and (ii) One-stage detectors (e.g., SSD, YOLO, RetinaNet).

Two-stage deep object detectors follow the classical approach of performing region selection first and then classifying the selected regions into objects. For region selection, modern detectors now use a so-called Region Proposal Network (RPN) which estimates an objectness and bounding box coordinates for each cell in a regular grid with respect to a set of fixed rectangular regions (called anchor boxes). The second stage then classifies each such region into object categories, also finetuning the bounding box coordinates in the meantime. 

Single-stage detectors, on the other hand, combine the two stages and estimate object class scores and bounding box coordinates simultaneously. In such detectors, similar to the RPN network in two-stage detectors, a fixed set of box configurations are assumed for each location and each class. This then means classifying a large number of boxes (approximately 8732 for SSD \cite{RETINANET}) per class.

\begin{table*}
\caption{A summary of our contributions in the context of relevant studies. Symbol ``x'' represents the number of parameters of the baseline network. NA means ``not applicable''. HD, H, A are short for horizontal Disparity, Height Above Ground and Angle With Gravity. UCM means Ultrametric Contour Map. (+) means separate network input. D$^{est}$ indicated that the model is also tested with estimated depth maps. 
}
\label{table:Comparison}
\centering
\begin{adjustbox}{width=\textwidth}
\small
\begin{tabular}{ | c | c c c c c c c c c |}

\hline
 & \CStart{} Base \\ Network\CEnd{} & \CStart{}Backbone  \\ Network\CEnd{} & \CStart{}Model \\ Parameter\CEnd{} & \CStart{} Model  \\ Input Type\CEnd{} &
 \CStart{}Concatenation Type\\  (Before/After RPN)\CEnd{} & \CStart{} Tested RGB \\Datasets\CEnd{} &  \CStart{}RGB Datasets \\ mAP Improvement\CEnd{} & \CStart{}Tested RGB-D \\ Datasets\CEnd{} & \CStart{}RGB-D Datasets \\ mAP Improvement\CEnd{}  \\
\hline
\hline
\cite{cao2017exploiting} (D$^{est}$) & \CStart{}Fast \\ R-CNN\CEnd{} & VGG-16 & 2x &  RGB + Gray-scale Depth Map  &  After & \CStart{}VOC 2007\CEnd{} & \checkmark & \CStart{}NYUD2 \\ B3DO\CEnd{}  & \checkmark  \\
\hline
\cite{gupta2014learning} &  \CStart{}R-CNN\CEnd{} & VGG-16 & 2x &  RGB + HDHA & After  & None & NA & NYUD2  & \checkmark  \\
\hline
\cite{hou2018object} &  R-CNN & AlexNet & $\geq$ 5x &  RGB + HD + H + A + UCM  & Before & None & NA & \CStart{}NYUD2 \\ SUN RGB-D\CEnd{}  & \checkmark  \\ \hdashline

\cite{hou2018object} & \CStart{}Fast \\ R-CNN\CEnd{} & VGG-16 & $\geq$ 4x &  RGB + HD + H + A  & Before & None & NA & NYUD2  & \checkmark  \\
\hline
\hline
Ours (D$^{est}$) &  \CStart{}Faster \\ R-CNN\CEnd{} & \CStart{}VGG-16 \\ or \\ ResNet-101\CEnd{} & $\leq$ 2x &  RGB + (Depth or HDHA)  & Before & VOC 2007 & \xmark & \CStart{} SUN RGB-D\CEnd{}  & \checkmark  \\ \hdashline
Ours (D$^{est}$) &  \CStart{}Faster \\ R-CNN\CEnd{} & \CStart{}VGG-16 \\ or \\ ResNet-101\CEnd{} & $\geq$ 4x &  RGB + (Depth or HDHA)  & After & VOC 2007 & \xmark & \CStart{} SUN RGB-D\CEnd{}  & \checkmark  \\
\hline
\end{tabular}
\end{adjustbox}
\end{table*} 

\subsection{Object Detection with Depth}


By definition, depth can be more informative than the corresponding visual image about the local surface structure of the objects as well as the layout of the scene. An object detector can utilize both cues for both determining the regions likely to contain objects as well as in classifying regions into objects. 

Although there are many studies using depth for 3D object detection \cite{shi2022stereo,ding2020learning,reading2021categorical, li2019stereo, liang2018deep, chen2017multi, liu2021ground}, only a few studies attempted integrating depth into deep object detectors \cite{cao2017exploiting,gupta2014learning,hou2018object} for 2D object detection. The work by Gupta et al. \cite{gupta2014learning} was one of the first to do so. They utilized the depth information for extending both the region proposal stage as well as the classification stage. To be specific, they used RGB-D images to compute 2.5D contours from which they estimated regions that are likely to contain objects. For the object classification stage, they first trained two CNN models for feature extraction; one for extracting features from the RGB information and one from the depth information. Then, they trained a linear SVM for classifying these features into objects. Another crucial contribution of Gupta et al. was to use horizontal disparity (D), height from ground (H) and angle of the surface normal with the direction of gravity (A) as the input to the feature extracting CNN model. They demonstrated that this performs better than providing depth directly.

It has also been demonstrated by Cao et al. \cite{cao2017exploiting} that depth estimated from an RGB image can be used to improve object detection from that RGB image, without relying on any depth sensor. They used Conditional Random Field to estimate the depth of the scene from the RGB image and train two independent CNN networks to classify the regions using the RGB inputs and the estimated depth input. For encoding the depth information, they directly provided the logarithm of the estimated depth and refrained from using surface normals or D, H and A encoding used by \cite{gupta2014learning}, claiming that such cues are not informative since (i) the estimated depth is approximate and noisy, and (ii) information about the camera is required for some of these cues.


Hou et al. \cite{hou2018object} provided a very informative analysis on the different mechanisms for integrating depth information into a deep object detector. Namely, they investigated the importance of the different visual inputs (RGB, depth, angle, height, contour, etc.) as well as the different levels (input-level vs. feature-level) for combining RGB and depth information for object detection. They argued and demonstrated that (i) color and depth should not be combined at the input level since they carry different types of information, and (ii) processing D, H and A separately with separate networks performs better than processing DHA together with a single network.

\subsection{Summary of the Literature and Our Contributions}


In the context of  related previous work, we make the following major contributions (see also Table \ref{table:Comparison} for a summary):
\begin{itemize}
    \item We extend Faster R-CNN \cite{FasterRCNN} with depth estimated from the input RGB image. To the best of our knowledge, ours is to first to perform such an extension on Faster R-CNN. This is important because Faster R-CNN is more sophisticated and capable than R-CNN or Fast R-CNN, which are used in \cite{cao2017exploiting,gupta2014learning,hou2018object}.
    \item Our extension suggests increasing the number of parameters twice in some configurations and even less in other configurations. Whereas these factors are more than four for \cite{hou2018object}, and two for \cite{cao2017exploiting,gupta2014learning} (Table \ref{table:Comparison}).
    \item We also provide results and improvement on an RGB dataset.
\end{itemize}

\renewcommand\theadalign{c}
\renewcommand\theadgape{\Gape[3pt]}
\renewcommand\cellgape{\Gape[3pt]}

\section{METHOD}
\label{sect:overview}
Object detection pertains to estimating object classes and positions given an input image, $I\in \mathbb{R}^{W\times H\times 3}$. As illustrated in Figure \ref{fig:Architecture}, our method is composed of three main steps:
\begin{itemize}
    \item Depth estimation (Section \ref{sect:depth_estimation}), which estimates a map, $D\in \mathbb{R}^{W\times H\times 1}$, containing depth, $d_i$, value for each pixel $i\in I$.
    \item Extracting and fusing features from RGB image ($I$) and estimated depth ($D$ -- Section \ref{sect:extraction_and_fusion}). We use a deep feature extractor, $\phi(\cdot)$, to extract $\phi(I)$ and $\phi(D)$.
    \item Object detection from the fused RGB and depth features ($\phi(I)$ and $\phi(D)$ -- Section \ref{sect:object_detection_from_depth}).
\end{itemize}

\subsection{Depth Estimation Network (DEN)}
\label{sect:depth_estimation}

Similar to the approach of Cao et al. \cite{cao2017exploiting}, we estimate depth ($D$) from a single RGB image ($I$) and aim to improve object detection with the estimated depth. For estimating the depth, we adapt and extend the deep network proposed by Hu et al. \cite{hu2019revisiting}. 

We train DEN separately from the object detector, on different datasets and their combinations (as explained in Appendix A.4 and Table  \ref{table:VOC2007TrainingSet}).  These datasets are as follows: NYU-D2 dataset \cite{silberman2012indoor} is used as the indoor scenes and Make3D-2 \cite{saxena2006learning,saxena2007learning} and KITTI  datasets \cite{geiger2013vision} as the outdoor scenes. 

\subsection{Encoding Depth}
\label{section:EncodingDepth}

Similar to color, depth can be encoded in different ways, which significantly affect the overall performance when they are used (see, e.g. \cite{gupta2014learning}). In this section, we describe the widely used depth encoding schemes which we investigate in the paper -- see also Figure \ref{fig:InputTypes} for how these encodings reflect different aspects of the 3D structure.


\subsubsection{Gray-scale Encoding}

In this encoding, the depth values are converted to gray-scale intensity values linearly as follows:
\begin{equation}
    g(d) = 255 \frac{d-d_\mathrm{min}}{d_\mathrm{max}-d_\mathrm{min}},
\end{equation}
where $d_\mathrm{{min}}$, $d_\mathrm{{max}}$ refer to the minimum and maximum depth values of all the scenes in the dataset.



\subsubsection{Jet Color Space Encoding}

It is known that encoding depth in the jet color space yields better performance for many 3D object recognition tasks \cite{eitel2015multimodal,song2015sun}. This is expected since the jet color space provides a bigger range for the depth values and makes depth discontinuities more distinctive (see also Figure \ref{fig:InputTypes}). For conversion into the jet color space, the grayscale encoding values are used as indices for a jet colormap with 256 entries.




\subsubsection{Horizonzal Disparity Height Angle (HDHA) Encoding} 

Recently, it has been shown that explicitly encoding horizontal disparity ($HD$), height (with respect to the ground -- denoted with $H$) and angle with the vertical direction ($A$) yields better depth representations \cite{gupta2014learning}. HDHA encoding, also called HHA or DHA in the literature, can be formally defined as follows: 
\begin{equation}
    HD = p_x - p^c_x,
\end{equation}
where $p_x$ is the $x$ coordinate of pixel $p$, and $p^c$ is the corresponding pixel  in camera $c$. 

As for $H$ (the height from the ground), since it is tricky to obtain the knowledge about the ground, the height from the lowest point in the scene is usually taken \cite{gupta2013perceptual}. 

To compute angle with gravity ($A$), an iterative procedure is used as in \cite{gupta2013perceptual}: An initial estimate for the gravity direction is taken as the vertical axis, with respect to which all surface normals are clustered into surfaces that are approximately parallel or orthogonal to the gravity direction. After clustering, a new gravity direction estimated with respect to the parallel and orthogonal clusters. These steps are iterated to minimize so that the gravity direction is as parallel as possible to the parallel surfaces and as orthogonal as possible to the orthogonal surfaces.

Before calculating the HDHA encoding, the grayscale depth encoding is zero-centered and normalized with standard deviation.

\subsection{Extracting and Fusing Features from RGB and Depth Data}
\label{sect:extraction_and_fusion}

The RGB input ($I$) and the estimated depth encoding ($HD$) are separately fed to a feature extraction network, yielding two sets of features, $\phi(I)\in \mathbb{R}^{W_2\times H_2 \times N_2}$ and $\phi(D)\in \mathbb{R}^{W_2\times H_2 \times N_2}$. The concatenation $\phi(I) \oplus \phi(D)$ is passed through a convolutional layer to reduce dimensions to $W_2\times H_2 \times N_2$.


\subsection{Object Detection from Fused RGB-Depth Features}
\label{sect:object_detection_from_depth}

In this work, we use Faster R-CNN \cite{FasterRCNN}, which is one of the state-of-the-art two-stage object detectors; however, our contributions are not specific to Faster R-CNN and they can be integrated into any object detector that are based on features extracted from a backbone network. We use the region proposal network and the bounding box regression \& classification stages as they are (except for the alternative architectures in the ablation experiments), and therefore, we only briefly remind these stages here.

In Faster R-CNN, the first stage, Region Proposal Network, estimates regions of interest with respect to a fixed set of bounding box configurations (called anchor boxes). The regions passing this first stage are then classified into object categories and their bounding boxes are fine-tuned with respect to the region-pooled features.

\begin{figure}[h]
\centering
{
\scriptsize
\renewcommand{\arraystretch}{0.3}
\newcommand{\SKALA}[0]{0.08}
\begin{tabular}{ m{1.5cm}  m{1.2cm} m{1.2cm} m{1.2cm} m{1.2cm}}
\CStart{}Input\CEnd{}         & 
    \includegraphics[scale=\SKALA]{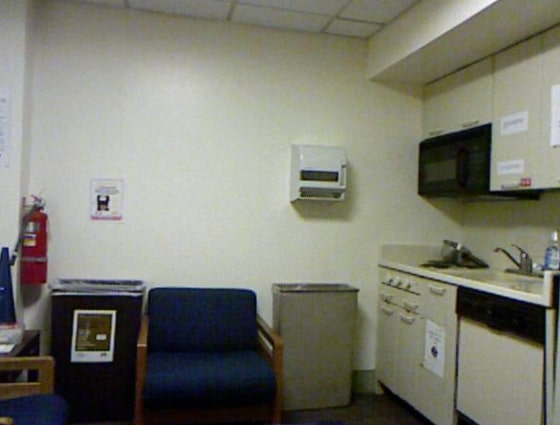} &
    \includegraphics[scale=\SKALA]{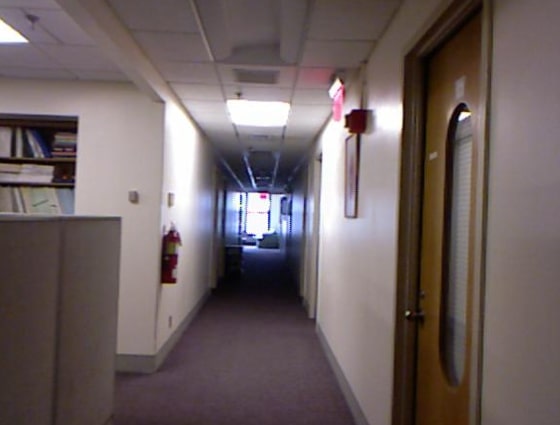} & 
    \includegraphics[scale=\SKALA]{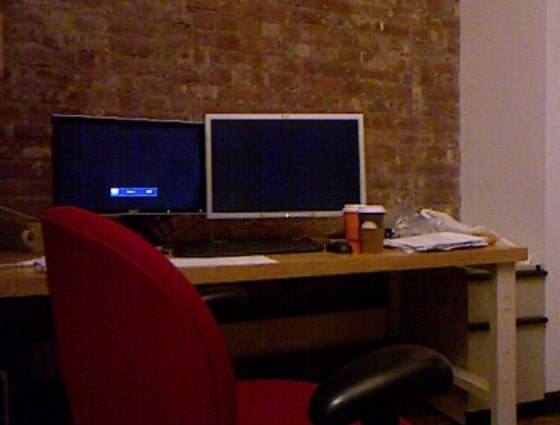} & 
    \includegraphics[scale=\SKALA]{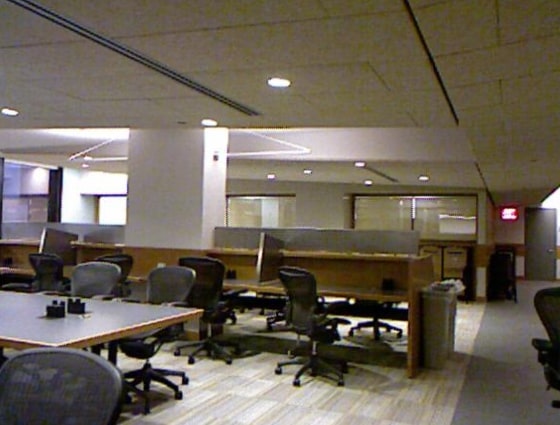} \\\\
\CStart{}Grayscale (GT)\CEnd{}         & 
    \includegraphics[scale=\SKALA]{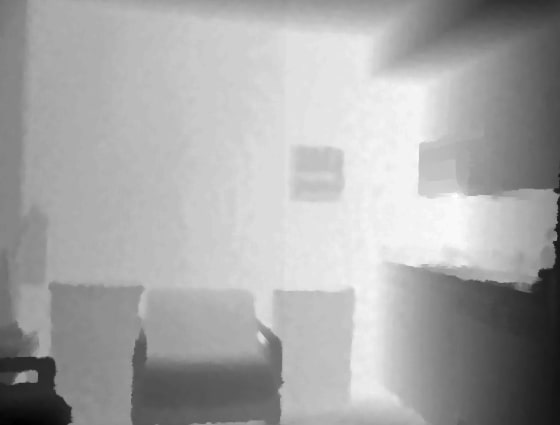} &
    \includegraphics[scale=\SKALA]{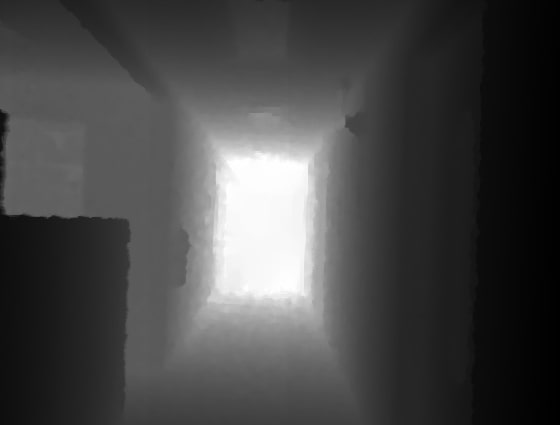} & 
    \includegraphics[scale=\SKALA]{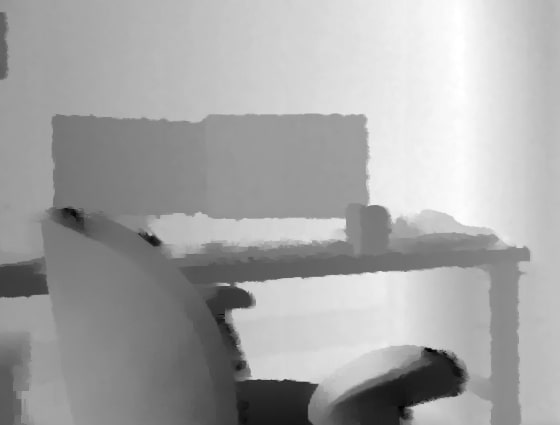} &
    \includegraphics[scale=\SKALA]{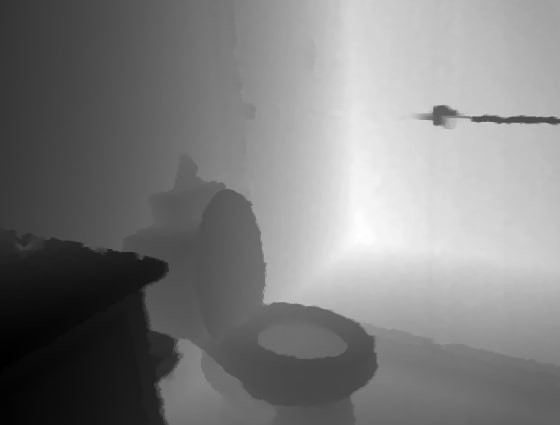} \\\\
\CStart{}Grayscale \\(Estimated)\CEnd{}         & 
    \includegraphics[scale=\SKALA]{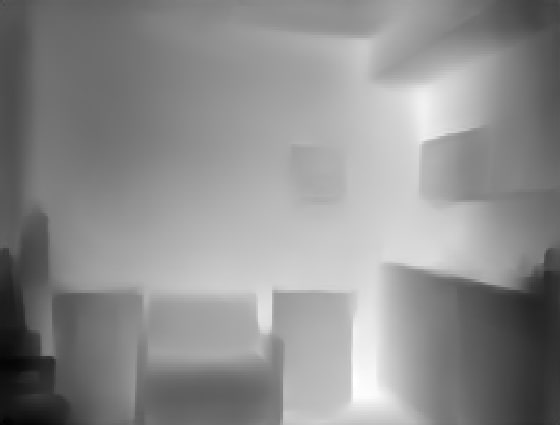} &
    \includegraphics[scale=\SKALA]{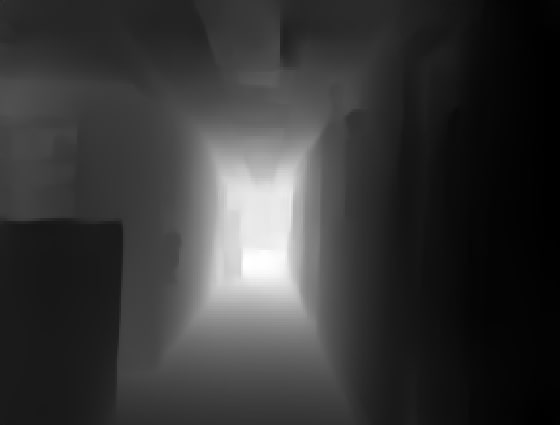} & 
    \includegraphics[scale=\SKALA]{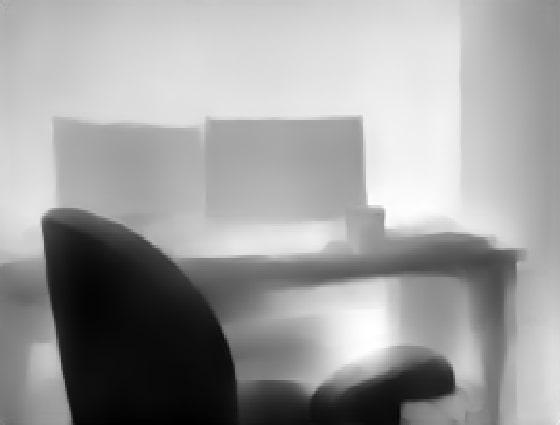} &
    \includegraphics[scale=\SKALA]{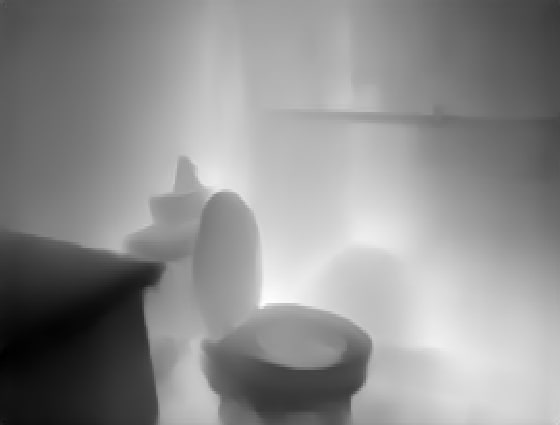} \\\\
\CStart{}Jet-scale (GT)\CEnd{}         & 
    \includegraphics[scale=\SKALA]{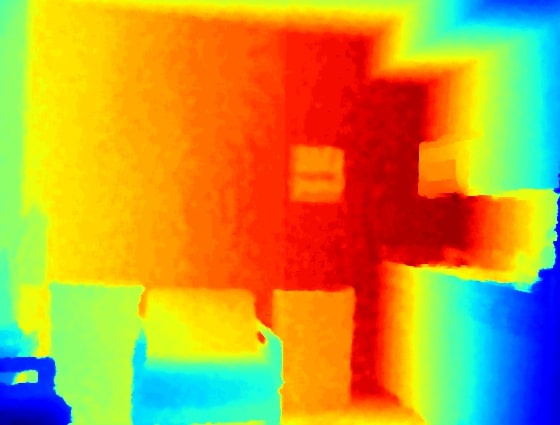} &
    \includegraphics[scale=\SKALA]{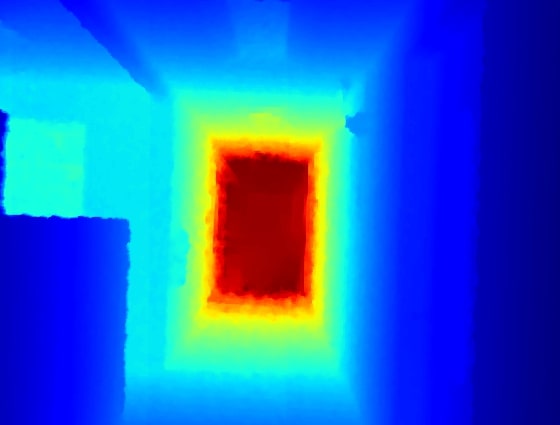} & 
    \includegraphics[scale=\SKALA]{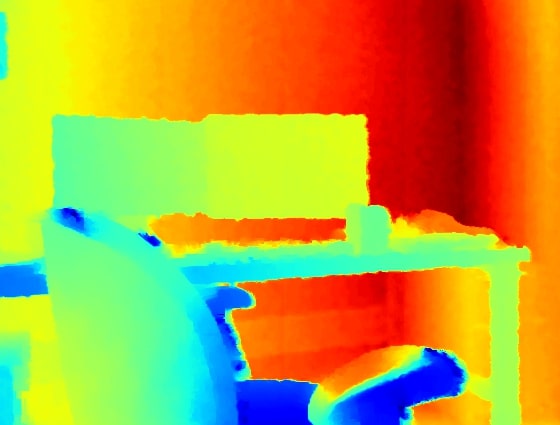} &
    \includegraphics[scale=\SKALA]{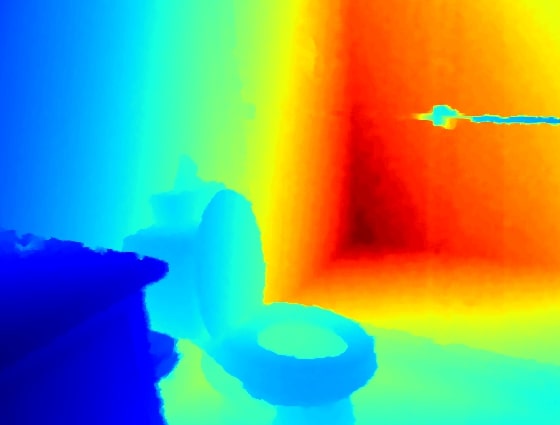} \\\\
\CStart{}Jet-scale \\(Estimated)\CEnd{}         & 
    \includegraphics[scale=\SKALA]{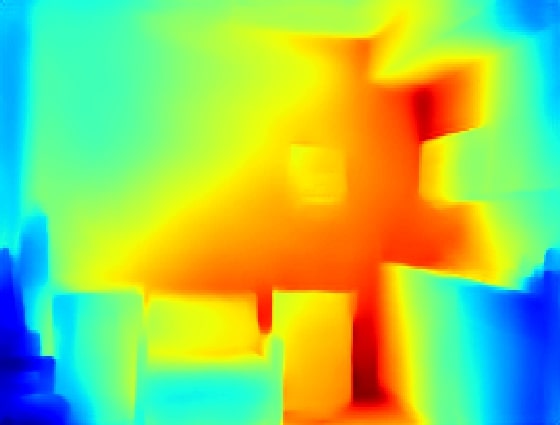} &
    \includegraphics[scale=\SKALA]{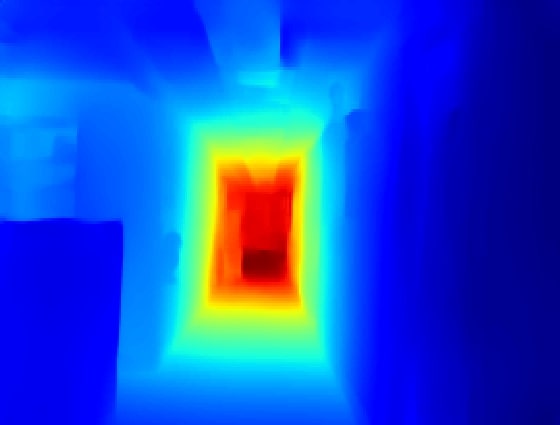} & 
    \includegraphics[scale=\SKALA]{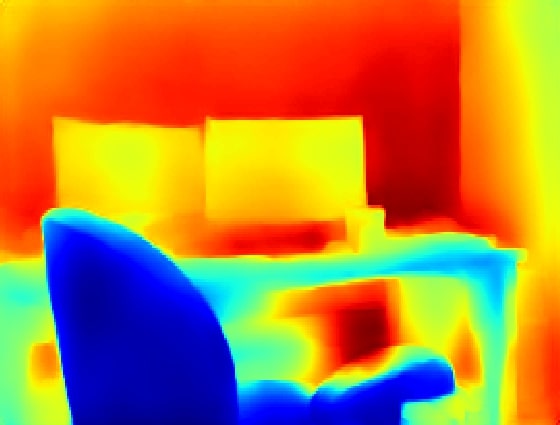} &
    \includegraphics[scale=\SKALA]{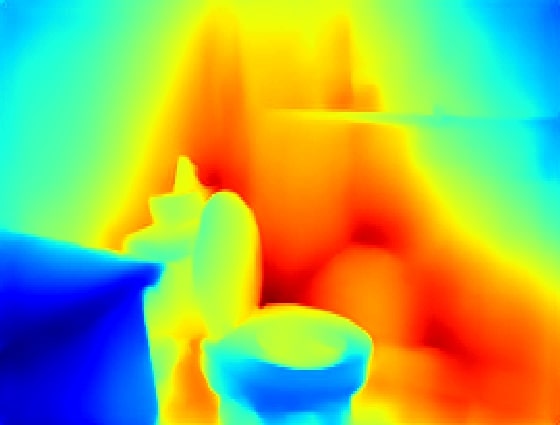} \\\\
\CStart{}HDHA (GT)\CEnd{}         & 
    \includegraphics[scale=\SKALA]{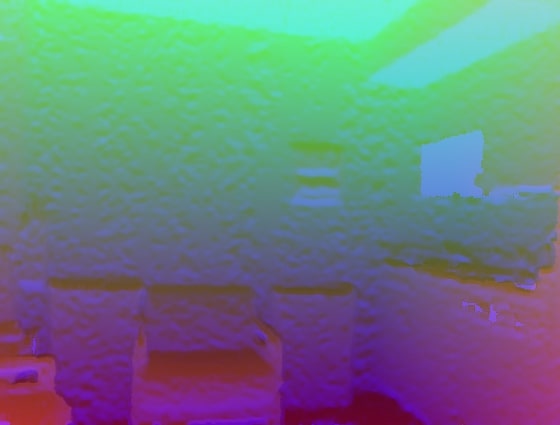} &
    \includegraphics[scale=\SKALA]{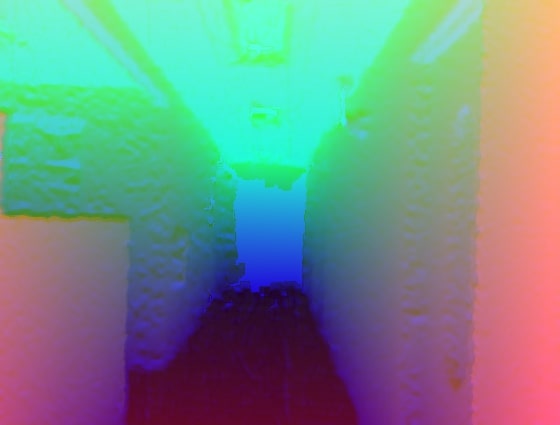} & 
    \includegraphics[scale=\SKALA]{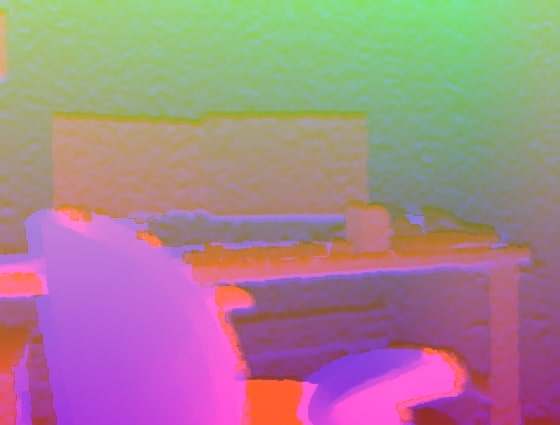} &
    \includegraphics[scale=\SKALA]{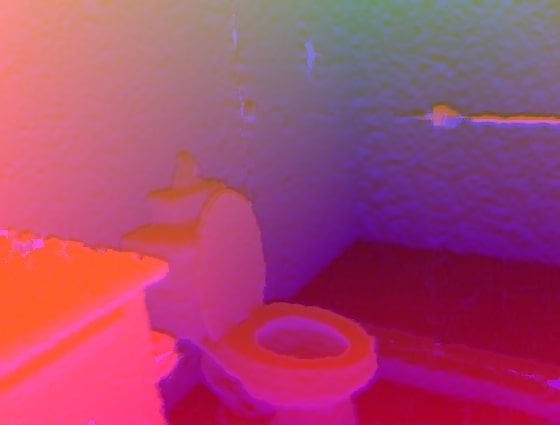} \\\\
\CStart{}HDHA\\(Estimated)\CEnd{}         & 
    \includegraphics[scale=\SKALA]{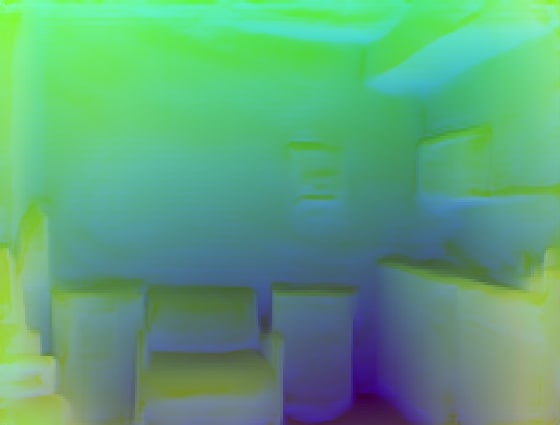} &
    \includegraphics[scale=\SKALA]{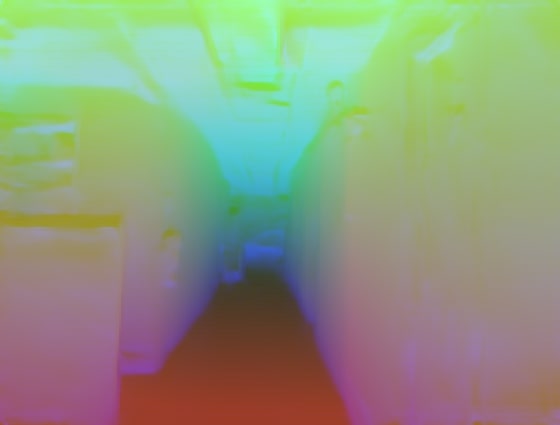} & 
    \includegraphics[scale=\SKALA]{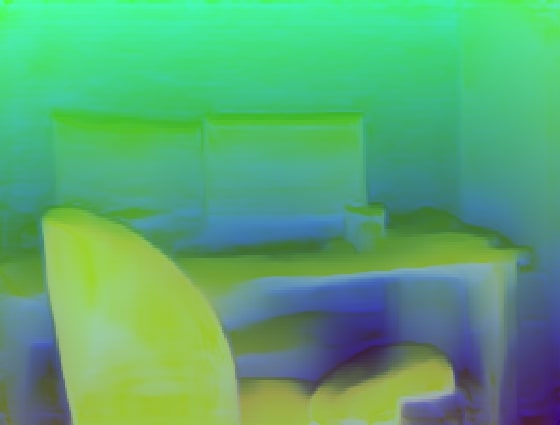} &
    \includegraphics[scale=\SKALA]{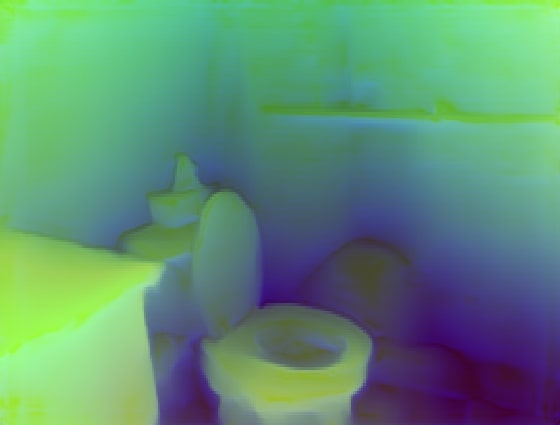} \\\\
\end{tabular}
}
\caption{The encoding schemes considered in this paper. Input images and ground truth depth maps are taken from the NYUD2 dataset \cite{silberman2012indoor}. \label{fig:InputTypes}}
\end{figure}




\begin{figure*}[h]
  \includegraphics[width=\linewidth]{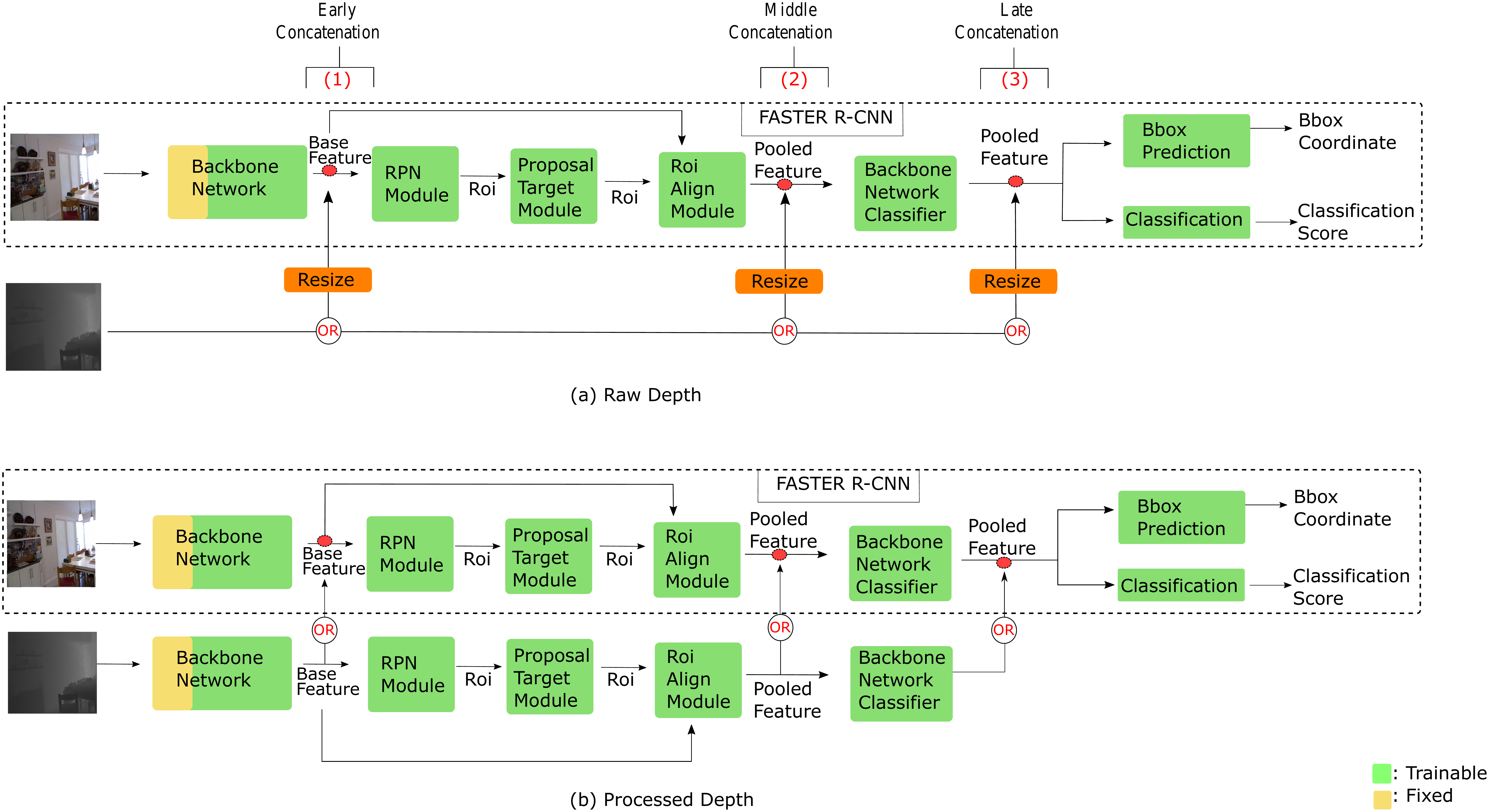}
  \caption{The alternative architectures evaluated in the paper. (a) Raw depth encoding integration. (b) Processed Depth integration.}
  \label{fig:All_Architecture}
\end{figure*}

\subsection{Alternative Architectures}

In addition to the architecture outlined in Figure \ref{fig:Architecture}, we also considered and extensively evaluated the following alternative architectures (see Figure \ref{fig:All_Architecture}): 
\begin{itemize}  
    \item[(1)] Early Concatenation: Integrating depth information before RPN, after having extracted features from the RGB image and the estimated depth.
    \item[(2)] Middle-Late Concatenation: Integrating depth information after the Roi Align Module. 
    \item[(3)] Late Concatenation: Integrating depth information just before the bounding box classification \& regression networks.
\end{itemize}

Also, we tested each architecture in two ways:
\begin{enumerate}
            \item Directly providing depth encoding as it is, without any feature extraction, shown in the upper part of Figure \ref{fig:All_Architecture}, (a) Raw Depth.
            \item Providing extracted features from depth, shown in bottom part of Figure \ref{fig:All_Architecture}(b).
\end{enumerate}

For details of integrating depth to these architectures, please refer to Section S2.

\section{Experiments and Results}
\label{sect:experiments}
In this section, we present our experimental results organized in four parts:
\begin{enumerate}[a)]
    \item \textbf{Architecture Experiments:} We explored three different ways of integrating the depth map into the object detection pipeline. Specifically, we tried adding the raw and processed depth information at various stages in processing. 
    \item \textbf{Depth Estimation Network Experiments:} We used different state-of-the-art depth estimator models to obtain the best depth maps.
    \item \textbf{Depth Estimation Training Set Experiments:} We trained the chosen depth estimator network with different combinations of indoor and outdoor depth datasets and observed how the training set affects object detection results.
    \item \textbf{Experiments for Object Detection:} Based on the best settings we obtained in the parts (a), (b) and (c), we evaluated object detection results for different datasets.
\end{enumerate}

\begin{table*}[h]
\caption{Object detection results on SUN RGB-D \MakeTextLowercase{(a)} and Pascal VOC 2007 \MakeTextLowercase{(b)} datasets for Architecture Experiments. The base coloumn is results of Faster R-CNN with using only RGB images. Training and test sets are same with official splits. EC,MC and LC represent initial concatenation, early concatenation, middle-late concatenation and late concatenation.
}
\label{table:AllArch}
\begin{subtable}{1\textwidth}
\centering
\begin{adjustbox}{width=\textwidth}
\small
\begin{tabular}{|c|c|cccccccccccccccccccc|c|}
\hline
   & & & \rotatebox{90}{bathtub} & \rotatebox{90}{bed} & \rotatebox{90}{b.shelf} & \rotatebox{90}{box} & \rotatebox{90}{chair} & \rotatebox{90}{counter} & \rotatebox{90}{desk} & \rotatebox{90}{door} & \rotatebox{90}{dresser} & \rotatebox{90}{g.bin} & \rotatebox{90}{lamp} & \rotatebox{90}{monitor} & \rotatebox{90}{n.stand} & \rotatebox{90}{pillow} & \rotatebox{90}{sink} & \rotatebox{90}{sofa} & \rotatebox{90}{table} & \rotatebox{90}{t.vision} & \rotatebox{90}{toilet} & \rotatebox{90}{mAP}\hfil \\ \hline
\hline
& Base & & 56.8	& 	80.1	& 	46.5	& 	15.4	& 	63.4	& 	51.9	& 	\textbf{30.3}	& 	55.8	& 	43.8	& 	\textbf{61.0}	& 	59.2	& 	44.2	& 	54.4	& 	54.0	& 69.1	& 	56.9	& 	\textbf{46.7}	& 	38.3 	& 	88.7	& 	53.5\\
 \hline
 \hline
 \multirow{4}{*}{Raw Depth}  & EC &  & 0.0 & 0.0 & 0.0 & 0.0 & 0.0 & 0.0 & 0.0 & 0.0 & 0.0 & 0.0 & 0.0  & 0.0 & 0.0 & 0.0 & 0.0 & 0.0 & 0.0 & 0.0 & 0.0 & 0.0 \\
\cline{2-23}
& MC &  & 0.0 & 0.0 & 0.0 & 0.0 & 0.0 & 0.0 & 0.0 & 0.0 & 0.0 & 0.0 & 0.0  & 0.0 & 0.0 & 0.0 & 0.0 & 0.0 & 0.0 & 0.0 & 0.0 & 0.0 \\
\cline{2-23}
& LC &  & 0.0 & 0.0 & 0.0 & 0.0 & 0.0 & 0.0 & 0.0 & 0.0 & 0.0 & 0.0 & 0.0  & 0.0 & 0.0 & 0.0 & 0.0 & 0.0 & 0.0 & 0.0 & 0.0 & 0.0 \\
\cline{1-23}
 \hline
 \multirow{3}{*}{Processed Depth} & EC &  & \textbf{60.4} & \textbf{81.4} & \textbf{51.7}	& \textbf{16.6} & \textbf{63.8} & \textbf{54.1} & 29.6 & \textbf{56.9} & \textbf{43.9}	& 59.0	& \textbf{62.5}	& \textbf{51.0} & \textbf{61.6} & \textbf{55.6} & \textbf{70.5}	& \textbf{59.7} & 46.1	& 36.5 & \textbf{91.5} & \textbf{55.4} \\
\cline{2-23}
& MC &  & 48.8 & 80.8 & 47.2 & 15.2	& 62.8 & 52.9 & 27.4 & 51.3 & 40.6 & 59.1	& 58.7	& 47.6	& 59.4 & 54.6 & 70.4 & 59.6 & 46.0 & \textbf{42.4} & 87.4 & 53.3  \\
\cline{2-23}
& LC &  & 49.4 & 80.3 & 48.3 & 16.1 & 62.2 & 51.3 & 26.9	& 51.1 & 40.3 & 59.0 & 57.5	&  49.6 & 58.3 & 53.3 & 66.7 & 58.6  & 45.5 & 39.3 & 84.5 & 52.5 \\
\hline 
\end{tabular}
\end{adjustbox}
\caption{SUN RGB-D}
\end{subtable}
\begin{subtable}{1\textwidth}
\centering
\begin{adjustbox}{width=\textwidth}
\small
\begin{tabular}{|c|c|cccccccccccccccccccc|c|}
\hline
   &  & \rotatebox{90}{A.plane} & \rotatebox{90}{Bicycle} & \rotatebox{90}{Bird} & \rotatebox{90}{Boat} & \rotatebox{90}{Bottle} & \rotatebox{90}{Bus} &  \rotatebox{90}{Car} &  \rotatebox{90}{Cat} &  \rotatebox{90}{Chair} &  \rotatebox{90}{Cow} &  \rotatebox{90}{D.table} &  \rotatebox{90}{Dog} &  \rotatebox{90}{Horse} &  \rotatebox{90}{M.bike} &  \rotatebox{90}{Person} &  \rotatebox{90}{P.plant} &  \rotatebox{90}{Sheep} &  \rotatebox{90}{Sofa} &  \rotatebox{90}{Train} &  \rotatebox{90}{Tv/mon.}&  \rotatebox{90}{mAP} \\ \hline
\hline
& Base & 76.3 & \textbf{81.8} & \textbf{77.0} & \textbf{66.2} & \textbf{60.6} & 80.2 & 85.5 & \textbf{86.5} & \textbf{55.6} & \textbf{83.3} & 65.6 & \textbf{85.8} & \textbf{85.3} & 77.9 & 78.6 & \textbf{48.2} & \textbf{75.9} & \textbf{74.2} & 77.9 & \textbf{76.2} & \textbf{74.9}   \\
 \hline
 \hline
 \multirow{4}{*}{Raw Depth}  & EC &  0.0 & 0.0 & 0.0 & 0.0 & 0.0 & 0.0 & 0.0 & 0.0 & 0.0 & 0.0 & 0.0 & 0.0  & 0.0 & 0.0 & 0.0 & 0.0 & 0.0 & 0.0 & 0.0 & 0.0 & 0.0 \\
\cline{2-23}
& MC & 0.0 & 0.0 & 0.0 & 0.0 & 0.0 & 0.0 & 0.0 & 0.0 & 0.0 & 0.0 & 0.0 & 0.0  & 0.0 & 0.0 & 0.0 & 0.0 & 0.0 & 0.0 & 0.0 & 0.0 & 0.0 \\
\cline{2-23}
& LC &  0.0 & 0.0 & 0.0 & 0.0 & 0.0 & 0.0 & 0.0 &0.0 & 0.0 & 0.0 & 0.0 & 0.0  & 0.0 & 0.0 & 0.0 & 0.0 & 0.0 & 0.0 & 0.0 & 0.0 & 0.0 \\
\cline{1-23}
 \hline
 \multirow{3}{*}{Processed Depth} & EC & \textbf{78.3} & 79.8 & \textbf{77.0} & 63.6 & 59.5 & \textbf{82.9} & \textbf{85.9} & 85.8 & 55.3 & 76.4 & 66.0 & 84.4 & 83.8 & 77.7 & \textbf{78.7} & 47.2 & 75.5 & 73.5 & \textbf{79.8} & 73.5 & 74.2   \\
\cline{2-23}
& MC &  77.5 & 79.3 & 75.8 & 62.3 & 59.1 & 81.3 & 83.8 & 83.9 & 55.3 & 82.1 & \textbf{66.1} & 82.1  & 83.7 & 77.3 & 78.1 & 48.1 & 73.8 & 72.8 & 79.4 & 73.1 & 73.7\\
\cline{2-23}
& LC & 73.9 & 79.7 & 75.9 & 60.4 & 58.2 & 80 & 82.7 & 84.4 & 50 & 79.9 & 62.6 & 82.8 & 81.7 & \textbf{79.4} & 77.2 & 45.9 & 72.1 & 73.5 & 76.7 & 72.7 & 72.5  \\
\hline
\end{tabular}
\end{adjustbox}
\caption{PASCAL VOC 2007}
\end{subtable}
\end{table*}

\subsection{Architecture Experiments}
\label{section:ArchExp}

In this section, we investigate the following two questions, with the ultimate goal of improving object detection accuracy: 

\textit{Q1: Should depth be used as is (i.e. as raw depth) or processed first before integrating into the pipeline?}
    
To answer this question, we identified three potential key concatenation points shown as red circles in Figure \ref{fig:All_Architecture}: namely, the output of the backbone network, the output of the RoI Align Module and the output of the RoI Feature Extractor module. We name these points respectively as Early Concatenation, Middle-Late Concatenation and Late Concatenation points. Since the answer to question Q1 does not depend on how the depth map was obtained, we used the estimated depth to make the comparison in SUN RGB-D and Pascal VOC datasets.

For this, first, we experimented with integrating the raw depth by combining the pooled RGB image features (of size N$\times$4096 where N is number of predicted objects), with estimated raw depth maps, without any feature extraction step. For more concatenation details, please refer to Section S2.

This concatenation yields a feature vector of size N$\times$8192.  For this case, shown in Figure \ref{fig:All_Architecture} case (a) Raw Depth and Late Concatenation, the training did not converge; as a result, mAP of Faster R-CNN was 0, shown in the fourth row of Table \ref{table:AllArch}.

Then, we tested Early and Middle-Late Concatenation points for raw depth map. Like Late Concatenation, the training did not converge for both concatenation points.

Therefore, we conclude that using the depth map in its raw form (without any feature extraction) harms the performance of Faster R-CNN dramatically.

\textit{Q2: What is the best stage to integrate depth into the object detection pipeline?}

Then, we tested processed depth maps with the three concatenation points displayed in Figure \ref{fig:All_Architecture}. First of all, we tested an earlier concatenation point where we integrate depth features into the object detection pipeline, as shown in Figure \ref{fig:All_Architecture} case (b), Early Concatenation (1). In this concatenation type, the tested architecture has two backbone networks and outputs of these backbone networks are fused. Then, the RPN module is fed with these fused features. Object detection results of this concatenation are shown in the fifth row of Table \ref{table:AllArch}, where we see that, on SUN RGB-D, early concatenation improves compared to the baseline (55.4 mAP vs. 53.5 mAP) whereas it is slightly worse on Pascal VOC (74.2 mAP vs. 74.9 mAP).

In Middle-Late Concatenation, each stream has its own backbone and region proposal network. To concatenate depth and RGB stream, outputs of RoI Align Modules are joined. RoI features are obtained from these joint features. Object detection results of this concatenation is shown in the sixth row of Table \ref{table:AllArch}, where we see that Middle-Late Concatenation worsens the detection performance.

In Late Concatenation, the tested architecture has nearly two complete object detection streams. Outputs of the two streams are concatenated just before the classification and bounding box regression tasks. Then these tasks are performed on concatenated features. Table \ref{table:AllArch} shows that this concatenation point provides the worst object detection results among the tested concatenation points.

The details of concatenation processed depth maps are presented Section S2.

These results suggest that processing the depth map like an RGB image appears to perform better than using it in its raw form. Previous works \cite{gupta2014learning, hou2018object, cao2017exploiting} also support this idea that depth map should be processed just like the RGB image (from the backbone network to the RoI feature extraction module) as illustrated in Figure \ref{fig:All_Architecture} case (b),  using Proc. Depth + Late Concatenation (3). The disadvantage of such methods is that the total number of network parameters increases significantly. In order to keep the number of parameters low and at the same time process depth using similar stages as in RGB processing, we propose to concatenate depth and RGB as early as possible and pass this combination through a common processing pipeline. Our proposal can be seen in Figure \ref{fig:All_Architecture} case (b), using Proc. Depth + Early (1), where depth map and RGB image separately pass through the backbone network and their output is concatenated just before a convolution layer. This convolution layer decreases the number of features and provides size matching with the pre-trained RoI Feature Extraction module. After the convolution layer, the rest of the object detection pipeline is the same as that of Faster R-CNN.

Using estimated depth map harms the Faster R-CNN performance for both Pascal VOC and SUN RGB-D datasets for Middle-Late and Late concatenation points, shown in Table \ref{table:AllArch}. While Early Concatenation point improves the Faster R-CNN performance for the SUN RGB-D dataset, it performs worse than the baseline for Pascal VOC.

\subsection{Experiments for the Depth Estimation Network and its Training Set}

We hypothesize that the quality of estimated depth maps could be a reason for this performance drop. The quality of the depth map depends on the depth estimation network, DEN, and the training set used for training DEN. Therefore, to analyze how estimated depth maps affect Faster R-CNN performance, we first try different state-of-the-art depth estimator networks, then we change its training sets. 

Initially, we used DEN proposed by Hu et al. \cite{hu2018revisiting} and we tested three more DENs which are \cite{watson2019self, li2018megadepth,lasinger2019towards}. We compared their results both qualitatively and quantitatively. While, the initially used DEN has the best estimated depth map for indoor scenes, the SUN RGB-D dataset, as shown in Figure S2, DEN proposed by Watson et al. \cite{watson2019self} has the best estimated depth map for outdoor scenes, i.e. the Pascal VOC dataset, as shown in Figure S1 and Table S1. Changing DENs slightly affects Faster R-CNN performance, as shown in Table A.1 and all of them decrease object detection performance of Faster R-CNN for the Pascal VOC dataset. Details of testing DENs are presented in Section S3.

While the DEN, proposed by Watson et al. \cite{watson2019self}, is trained with indoor and outdoor datasets, DEN proposed by Hu et al. \cite{hu2018revisiting} is only trained with indoor scenes. Therefore, we trained it with a different combination of indoor and outdoor datasets. For each combination, we tested Early Concatenation with Resnet-101 backbone and Pascal VOC dataset. However, changing the training set of DEN slightly affects Faster R-CNN performance, shown in Table \ref{table:VOC2007TrainingSet} and estimated depth maps do not improve the Faster R-CNN performance for any combination of training sets. Figure S3 shows visual outputs of tested training sets. Details of testing training set of DEN are presented in Section S4. 

While integrating estimated depth to Faster R-CNN degrades object detection performance for the Pascal VOC dataset, using ground-truth or estimated depth  improves it for the  SUN RGB-D dataset. Experiments with ground-truth depth map are presented in the Section \ref{sec:ExpObjDetect}.

\begin{table}[h]
\caption{Object detection results on Pascal VOC 2007 Results for depth   estimation network training with different datasets. The backbone network is ResNet-101 and the additional input type is estimated gray-scale depth map. Training set is VOC 2007 official trainval split and test set is VOC 2007  official test split.} 
\label{table:VOC2007TrainingSet}
\centering
\small
\begin{tabular}{|c|c|c|}
\hline
Training Set & \# of Data & mAP \\
\hline
\hline
Only RGB & - & \textbf{74.9} \\
\hline
\hline
NYU-D2  & \char`\~45k & 74.2 \\
\hline
NYU + Make3D & \char`\~1k & 74.1	 \\
\hline
KITTI & \char`\~24k &  73.2	 \\
\hline
KITTI+ NYU-D2 (a) & \char`\~70K  &  73.5  \\
\hline
KITTI+ NYU-D2 (b) & \char`\~70K  &  73.3  \\
\hline
\end{tabular}
\end{table} 



\subsection{Experiments for Object Detection}
\label{sec:ExpObjDetect}
After choosing the best architecture and the depth estimation network, we ran our method on different datasets with different ground-truth and estimated input types which are gray-scale depth map, depth map in jet color space and HDHA encoding (Figure \ref{fig:InputTypes}). We experimented with VGG-16 \cite{simonyan2014very} and ResNet-101 \cite{he2016deep} as backbone, and Faster-R CNN as base object detector. SUN RGB-D and Pascal VOC 2007 datasets are used.

\begin{table*}[h]
\caption{Object Detection Results on SUN RGB-D dataset. The first column shows  results of Faster R-CNN. The next three columns show ground-truth input types results. The last three columns show the estimated input types results. G-Depth means depth map in grayscale and J-Depth means depth map in jet color map. Training and test sets are same with official splits.
} 
\label{table:SUNRGB-D}
\centering
\begin{adjustbox}{width=1\textwidth}
\small
\begin{tabular}{|c|c|ccccccccccccccccccc|c|}
\hline
  & \rotatebox{90}{B.bone} & \rotatebox{90}{bathtub} & \rotatebox{90}{bed} & \rotatebox{90}{b.shelf} & \rotatebox{90}{box} & \rotatebox{90}{chair} & \rotatebox{90}{counter} & \rotatebox{90}{desk} & \rotatebox{90}{door} & \rotatebox{90}{dresser} & \rotatebox{90}{g.bin} & \rotatebox{90}{lamp} & \rotatebox{90}{monitor} & \rotatebox{90}{n.stand} & \rotatebox{90}{pillow} & \rotatebox{90}{sink} & \rotatebox{90}{sofa} & \rotatebox{90}{table} & \rotatebox{90}{t.vision} & \rotatebox{90}{toilet} & \rotatebox{90}{mAP}\hfil \\
 \hline
\hline
RGB & VGG-16 & 50.6 & 73.9 & 48.5 & 12.6 & 60.3 & 47.1 & 27.6 & 52.7 & 38.5 & 57.0 & 59.7 & 45.2 & 58.5 & 51.0 & 61.5 & 51.7 & 42.4 & 36.6 & 86.8 & 50.6   \\
\hline
RGB+HDHA* &VGG-16& \textbf{61.6} & \textbf{83.3} & 47.3 & \textbf{17.5} & 62.8 & 50.1 & 28.6 & \textbf{54.1} & \textbf{44.1} & 59.8 & 61.2 & \textbf{48.4} & 59.7 & \textbf{56.6} & \textbf{69.6} & \textbf{57.0} & 47.2 & 41.4 & 87.0 & \textbf{54.6} \\
\hline
RGB+GDep* &VGG-16& 59.7 & 81.5 & 47.6 & 16.4 & \textbf{64.0} & 49.4 & \textbf{29.6} & 53.0 & 42.2 & \textbf{60.2} & \textbf{61.8} & 47.5 & 61.7 & 55.7 & 64.7 & 56.0 & \textbf{48.0} & \textbf{45.7} & 84.8 & 54.2   \\
\hline
RGB +JDep* &VGG-16& 69.5 & 80.9 & 49.2 & 16.8 & 62.6 & \textbf{51.0} & 29.0 & 53.3 & 42.9 & 59.4 & 61.2 & 47.7 & \textbf{62.5} & 54.2 & 63.5 & 55.9 & 46.2 & 41.4 & 88.8 & 54.5    \\
\hline
RGB+HDHA &VGG-16& 56.6 & 78.0 & \textbf{50.4} & 15.4 & 60.8 & 49.0 & 27.7 & 53.6 & 42.6 & 58.0 & 58.5 & 44.7 & 60.6 & 49.3 & 63.5 & 55.5 & 46.1 & 38.0 & 85.4 & 52.3 \\
\hline
RGB+GDep &VGG-16& 56.6 & 77.7 & 44.4 & 14.2 & 60.4 & 48.9 & 25.6 & 51.8 & 41.3 & 57.2 & 58.9 & 44.9 & 59.7 & 50.8 & 62.2 & 52.6 & 45.6 & 38.6 & \textbf{88.9} & 51.6  \\
\hline
RGB+JDep &VGG-16& 60.8 & 78.9 & 47.6 & 15.0 & 61.6 & 48.8 & 27.3 & 53.6 & 41.0 & 56.6 & 58.3 & 43.7 & 58.5 & 49.3 & 64.3 & 55.7 & 45.1 & 36.8 & 85.4 & 52.0  \\
\hline
\hline
RGB & Res-101 & 56.8	& 	80.1	& 	46.5	& 	15.4	& 	63.4	& 	51.9	& 	30.3	& 	55.8	& 	43.8	& 	61.0	& 	59.2	& 	44.2	& 	54.4	& 	54.0	& 69.1	& 	56.9	& 	46.7	& 	38.3 	& 	88.7	& 	53.5\\
\hline
RGB+HDHA* & Res-101 & 68.3 & 85.4 & 52.1 & 19.4 & 66.7 & \textbf{54.4} & \textbf{33.4} & 58.5 & 43.4 & 58.8 & 61.5 & 49.7 & 60.4 & 60.1 & 73.6 & 62.5 & 46.4 & 47.3 & 90.6 & 57.5  \\
\hline
RGB+GDep* &Res-101 & \textbf{69.4} & 84.8 & 50.8 & \textbf{19.8} & \textbf{67.8} & 52.3 & 32.0 & 58.3 & 46.4 & 61.4 & \textbf{64.0} & \textbf{52.2} & \textbf{61.6} & \textbf{60.9} & \textbf{75.9} & \textbf{63.0} & \textbf{48.1} & \textbf{50.1} & 88.9 & \textbf{58.3}  \\
\hline
RGB+JDep* &Res-101 & 69.3 & \textbf{85.5} & \textbf{53.1} & 18.3 & 65.9 & 51.4 & 30.7 & \textbf{59.2} & \textbf{47.0} & \textbf{61.5} & 62.8 & 50.8 & 59.9 & 58.6 & 73.7 & 61.9 & 47.4 & 38.6 & 89.3 & 57.1  \\
\hline
RGB+HDHA &Res-101 & 57.5 & 81.8 & 49.8 & 15.8 & 64.5 & 51.4 & 26.9 & 57.4 & 43.3 & 58.8 & 61.7 & 48.3 & 60.5 & 56.8 & 70.5 & 58.7 & 44.3 & 41.9 & 88.8 & 54.7  \\
\hline
RGB+GDep &Res-101 & 60.4 & 81.4 & 51.7	& 16.6 & 63.8 & 54.1 & 29.6 & 56.9 & 43.9	& 59.0	& 62.5	& 51.0 & \textbf{61.6} & 55.6 & 70.5	& 59.7 & 46.1	& 36.5 & \textbf{91.5} & 55.1 \\
\hline
RGB+JDep &Res-101 & 58.9 & 80.5 & 50.3 & 16.3 & 64.0 & 53.5 & 29.8 & 56.8 & 46.7 & 58.4 & 62.2 & 49.0 & 59.1 & 56.0 & 72.9 & 59.6 & 47.3 & 41.1 & 86.2 & 55.2  \\
\hline
\end{tabular}
\end{adjustbox}
\end{table*}

\begin{table*}
\caption{Object Detection Results on Pascal VOC 2007 indoor categories. Training set is VOC 2007 official trainval split and test set is VOC 2007 official test split. The first row shows   results for Faster R-CNN. Other rows show additional estimated input types results. JDep means depth image in jet color space. GDep means gray-scale depth map.} 
\label{table:VOC2007Indoor}
\centering
\scriptsize
\begin{tabular}{|c|c|cccccc|c|}
\hline
  & \rotatebox{90}{B.Bone} & \rotatebox{90}{Bottle} & \rotatebox{90}{Chair} & \rotatebox{90}{D.Table} & \rotatebox{90}{P.Plant} & \rotatebox{90}{Sofa} & \rotatebox{90}{Tv/Mon.} &  \rotatebox{90}{mAP} \\
\hline\hline
RGB & VGG-16 & 54.8	& 	\textbf{53.5}	& 	65.1	& 43.6	& 	\textbf{69.4}	& 	75.1	& 	\textbf{60.3} \\
\hline
RGB+HDHA & VGG-16 & \textbf{57.3}	& 49.8	& 65.5	& 42.8	& 66.3	 & 76.0 & 59.6 	 \\
\hline
RGB+JDep & VGG-16 & 54.2	& 50.7 	& \textbf{67.1} 	& 39.6 	& 	68.9	& \textbf{76.6}	& 59.5 	 \\
\hline
RGB+GDep & VGG-16 & 53.1 & 50.4	& 65.4	& \textbf{43.8} 	& 	62.3 & 75.6		&  58.5 \\
\hline
\hline
RGB & Res-101 & 62.4	& 	59.5	& 	67.2	& 47.7	& 	\textbf{78.1}	& 	\textbf{78.2}	& 	\textbf{65.5} \\
\hline
RGB+HDHA & Res-101 & 63.0	& 	57.0	& 	\textbf{68.0}	& \textbf{49.8}	& 	75.2 & 	75.4	& 	64.7 \\
\hline
RGB+JDep & Res-101 & \textbf{63.2}	& 	59.4	& 	65.8	& 47.7	& 	74.1	& 	77.5	& 	64.6 \\
\hline
RGB+GDep & Res-101 &61.6	& 	\textbf{59.7}	& 	66.9	& 48.3	& 	77.1	& 	77.6	& 	65.2\\
\hline
\end{tabular}
\end{table*}

\begin{table*}[h]
\caption{ Object Detection Results on Pascal VOC 2007 dataset. Training set is VOC 2007 official trainval split and test set is VOC 2007  official test split.  The first row shows   results for Faster R-CNN. Other rows show additional estimated input types results. JDep means depth image in jet color space. GDep means gray-scale depth map.} 
\label{table:VOC2007}
\centering
\begin{adjustbox}{width=\textwidth}
\small
\begin{tabular}{|c|c|cccccccccccccccccccc|c|}
\hline
  & \rotatebox{90}{B.Bone} & \rotatebox{90}{A.plane} & \rotatebox{90}{Bicycle} & \rotatebox{90}{Bird} & \rotatebox{90}{Boat} & \rotatebox{90}{Bottle} & \rotatebox{90}{Bus} &  \rotatebox{90}{Car} &  \rotatebox{90}{Cat} &  \rotatebox{90}{Chair} &  \rotatebox{90}{Cow} &  \rotatebox{90}{D.table} &  \rotatebox{90}{Dog} &  \rotatebox{90}{Horse} &  \rotatebox{90}{M.bike} &  \rotatebox{90}{Person} &  \rotatebox{90}{P.plant} &  \rotatebox{90}{Sheep} &  \rotatebox{90}{Sofa} &  \rotatebox{90}{Train} &  \rotatebox{90}{Tv/mon.}&  \rotatebox{90}{mAP} \\
\hline\hline
RGB  & VGG-16 & 69.0 & 76.6 & 66.5 & 54.7 & 53.0 & 78.9 & \textbf{85.4} & \textbf{83.5} & \textbf{49.1} & \textbf{79.9} & \textbf{59.4} & 80.8 & 81.2 & 75.3 & 76.9 & \textbf{44.6} & 68.5 & 65.2 & 74.1 & 72.7 & \textbf{69.8}  \\
\hline
RGB+HDHA & VGG-16 & \textbf{71.0} & 77.7 & 64.9 & 54.9 & \textbf{54.9} & \textbf{79.1} & 84.5 & 82.9 & 48.6 & 74.8 & 62.9 & 78.3 & 81.1 & \textbf{76.0} & 76.7 & 44.4 & 69.2 & \textbf{66.7} & 73.5 & 70.7 & 69.6   \\
\hline
RGB+JDep & VGG-16 & 70.8 & \textbf{78.4} & 65.9 & \textbf{55.9} & 49.0 & 78.1 & 84.6 & 81.8 & 46.9 & 78.3 & 65.8 & 79.5 & \textbf{82.0} & 75.8 & 76.9 & 43.1 & \textbf{70.4} & 66.5 & 73.9 & 72.4 & \textbf{69.8}   \\
\hline
RGB+GDep & VGG-16 & 70.9 & 75.8 & \textbf{68.1} & 55.3 & 52.3 & 75.6 & 83.8 & 80.9 & 47.6 & 76.0 & 64.7 & \textbf{81.0} & 81.8 & 73.5 & \textbf{77.0} & 44.0 & 69.2 & 65.9 & \textbf{74.8} & \textbf{73.5} & 69.6  \\
\hline
\hline
RGB &  Res-101 & 76.3 & \textbf{81.8} & \textbf{77.0} & \textbf{66.2} & \textbf{60.6} & 80.2 & 85.5 & \textbf{86.5} & \textbf{55.6} & \textbf{83.3} & 65.6 & \textbf{85.8} & \textbf{85.3} & \textbf{77.9} & 78.6 & 48.2 & \textbf{75.9} & 74.2 & 77.9 & \textbf{76.2} & \textbf{74.9} \\
\hline
RGB+HDHA & Res-101 & \textbf{79.1} & 80.0 & 76.0 & 64.9 & 59.2 & \textbf{85.0} & 86.5 & 85.5 & 53.8 & 80.6 & 67.1 & 85.0 & 84.7 & 76.8 & \textbf{78.7} & \textbf{48.5} & 73.6 & 73.7 & 78.0 & 72.5 & 74.5   \\
\hline
RGB+JDep & Res-101 & 78.3 & 78.6 & 76.6 & 66.0 & 59.8 & 81.3 & \textbf{86.9} & 85.8 & 54.8 & 79.1 & \textbf{69.0} & 84.3 & 81.9 & 77.6 & 78.6 & 48.1 & 73.3 & \textbf{74.8} & \textbf{80.6} & 75.8 & 74.6 \\
\hline
RGB+GDep & Res-101 & 78.3 & 79.8 & \textbf{77.0} & 63.6 & 59.5 & 82.9 & 85.9 & 85.8 & 55.3 & 76.4 & 66.0 & 84.4 & 83.8 & 77.7 & \textbf{78.7} & 47.2 & 75.5 & 73.5 & 79.8 & 73.5 & 74.2 \\
\hline
\end{tabular}
\end{adjustbox}
\end{table*}

\begin{table}
\caption{Object Detection Results on the COCO dataset. Training set is the COCO 2017 official train split and test set is the COCO 2017 official validation split. The first row shows results for Faster R-CNN. GDep means gray-scale depth map.} 
\label{table:COCO}
\centering
\scriptsize
\begin{tabular}{|c|cccccc|}
\hline
  & AP & AP$_{50}$ & AP$_{75}$ & AP$_S$ & AP$_M$ & AP$_L$ \\
\hline\hline
RGB & \textbf{26.9} & \textbf{46.2} & \textbf{27.7} & \textbf{10.0} & \textbf{29.8} & \textbf{41.1} \\
RGB + GDep & 26.4 & 45.2 & 27.1 & 9.7 & 28.9 & 40.7\\
\hline
\end{tabular}
\end{table} 

Tables \ref{table:SUNRGB-D},  \ref{table:VOC2007Indoor}, \ref{table:VOC2007} and \ref{table:COCO} show object detection results in terms of mAP for SUN RGB-D, Pascal VOC indoor classes, PASCAL VOC all classes, and COCO, respectively. 

Note that previous work \cite{cao2017exploiting,gupta2014learning} used R-CNN \cite{FastRCNN} and Fast R-CNN \cite{girshick2015fast} respectively for the task; however, the idea behind using additional input is the same : Additional input and RGB images pass through separate networks until reaching the bounding box regression layer. Before passing through this layer, features obtained from these two separate networks are concatenated. Bounding box regression and classification tasks are performed on these fused features. This architecture is equivalent to Late Concatenation. For a fair comparison between our and previous methods, we implemented previous works’ idea on Faster R-CNN. Figure \ref{fig:All_Architecture}, case (b-3) shows this architecture in detail.

Tables \ref{table:PrevParam_1}, \ref{table:PrevmAP_HHA_1}  provide results in terms of mAP and the number of parameters. Our method outperforms the idea of late concatenation used in \cite{cao2017exploiting,gupta2014learning} in all experiments, as shown in Table \ref{table:PrevmAP_HHA_1}.

\begin{figure*}[hbt!]
  \includegraphics[width=\linewidth]{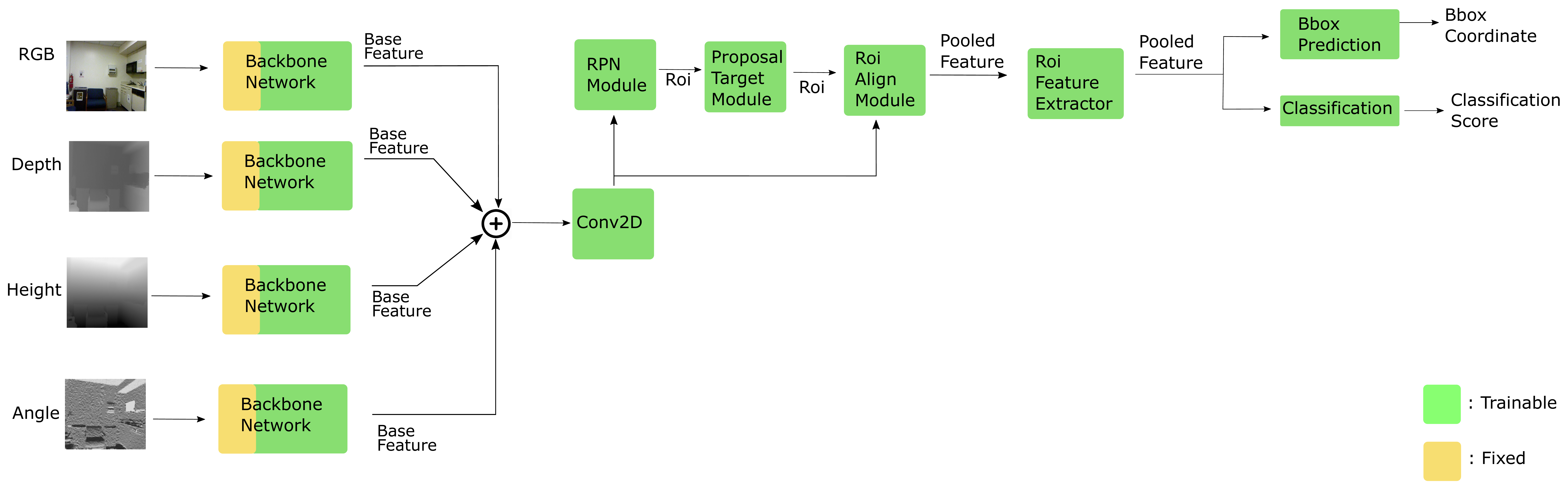}
  
  \caption{ Processing each channel of HDHA encoding separately proposed  in \cite{hou2018object} on Faster R-CNN.}
  \label{fig:PrevArchitecture}
\end{figure*}

Also, we evaluated Faster R-CNN by separately processing each channel of HDHA encoding, proposed in \cite{hou2018object}. Hou et al.'s method \cite{hou2018object} uses additional convolution layers just after the features concatenation of HDHA encoding channels and RGB images. Therefore, their method increases the base object detection model at least four times for the ResNet-101 backbone. However, we combined their idea, i.e. each channel of HDHA encoding has separate backbone network, and our architecture, i.e. using one convolution layer to decrease feature map channel just after the features concatenation.
Figure \ref{fig:PrevArchitecture} shows the pipeline of this idea and Tables \ref{table:PrevParam_1} and \ref{table:PrevmAP_HHA_1} show comparisons in terms of mAP and number of parameters. Although there is a huge number of parameters difference between our method and \cite{hou2018object}, our method performs better than all previous approaches \cite{hou2018object,cao2017exploiting,gupta2014learning}.

\begin{table*}[h]
\caption{A comparison of models in terms of the number of parameters. }
\label{table:PrevParam_1}
\centering
\scriptsize

\begin{tabular}{ | c |  c  c c c |c c c c | }

\hline
& \multicolumn{4}{c}{ VGG-16 } & \multicolumn{4}{|c|}{ ResNet101 }  \\
\hline \hline
 & Faster R-CNN & Ours &  \cite{cao2017exploiting,gupta2014learning} & \cite{hou2018object} & Faster R-CNN & Ours & \cite{cao2017exploiting,gupta2014learning} & \cite{hou2018object} \\
\hline
Trainable Parameters & 137M & 152M & 273M & 181M & 47M & 76M & 94M & 133M \\
\hline
Fixed Parameters & 260K & 520K & 520K & 1M & 328K & 632K & 655K & 1M \\
\hline
Total Parameters & 137M & 152M  &  274M & 182M & 47M & 77M  & 95M  & 135M\\
\hline
\end{tabular}
\end{table*}

\begin{table*}[h]
\caption{ Object detection performance (mAP) comparison between Faster R-CNN,  proposed  architecture, and previous works' architecture for the case of using HDHA encoding and Gray-Scale depth maps  as the additional input. For the sake of easier comparison, the results from Tables  \ref{table:SUNRGB-D}, \ref{table:VOC2007} are provided here again.} 
\label{table:PrevmAP_HHA_1}
\centering
\small
\begin{tabular}{|cc|ccc|cc|}
\hline
\multicolumn{2}{|c|}{Encoding Types} & Model  & \multicolumn{2}{|c}{ SUN RGB-D}  & \multicolumn{2}{|c|}{ Pascal VOC 2007} \\
\hline
 &  && VGG-16 & ResNet-101   & VGG-16 & ResNet-101  \\
\hline
\multicolumn{1}{|c|}{RGB} & - & Faster R-CNN  & 50.6 & 53.5 & 69.8 & 74.9 \\
\hline
\hline
 \multicolumn{1}{|c|} {\multirow{6}{*}{\CStart{}RGB \\+ \\HDHA\CEnd{}} }  & \multirow{3}{*}{\CStart{}Ground- \\Truth\CEnd{}}  &  \cite{hou2018object}  & 52.3 & 56.7 & - & - \\
\cline{3-7}
\multicolumn{1}{|c|}{}&& \cite{cao2017exploiting,gupta2014learning}  & 50.9 & 54.4 &- &- \\
\cline{3-7}
\multicolumn{1}{|c|}{}&& Ours & 54.6 & 57.5 &- &-\\
\cline{2-7}
\cline{2-7}
\multicolumn{1}{|c|}{}&\multirow{3}{*}{Estimated} &  \cite{hou2018object}  & 51.0 & 53.9 & 69.6 & 73.8 \\
\cline{3-7}
\multicolumn{1}{|c|}{}&& \cite{cao2017exploiting,gupta2014learning}  & 49.1 & 52.5 & 66.1 &71.7\\
\cline{3-7}
\multicolumn{1}{|c|}{}&& Ours  & 52.3 & 54.7 & 69.6 &  74.5\\
\hline
\hline
\multicolumn{1}{|c|}{\multirow{4}{*}{\CStart{}RGB \\+ \\Gray-Scale Depth\CEnd{}}}  & \multirow{2}{*}{\CStart{}Ground- \\Truth\CEnd{}}&\cite{cao2017exploiting,gupta2014learning} & 51.2 & 55.7 & -&- \\
\cline{3-7}
\multicolumn{1}{|c|}{}&& Ours  & 54.2 & 58.3 &- &-\\
\cline{2-7}
\cline{2-7}
\multicolumn{1}{|c|}{}& \multirow{2}{*}{Estimated}&  \cite{cao2017exploiting,gupta2014learning} &  48.3 & 52.5 & 66.4 & 72.5 \\
\cline{3-7}
\multicolumn{1}{|c|}{}&&Ours & 52.0 & 55.4 & 69.6 &  74.2 \\
\hline
\end{tabular}
\end{table*}

\begin{figure*}[hbt!]
   \centering
    \includegraphics[width=0.8\textwidth]{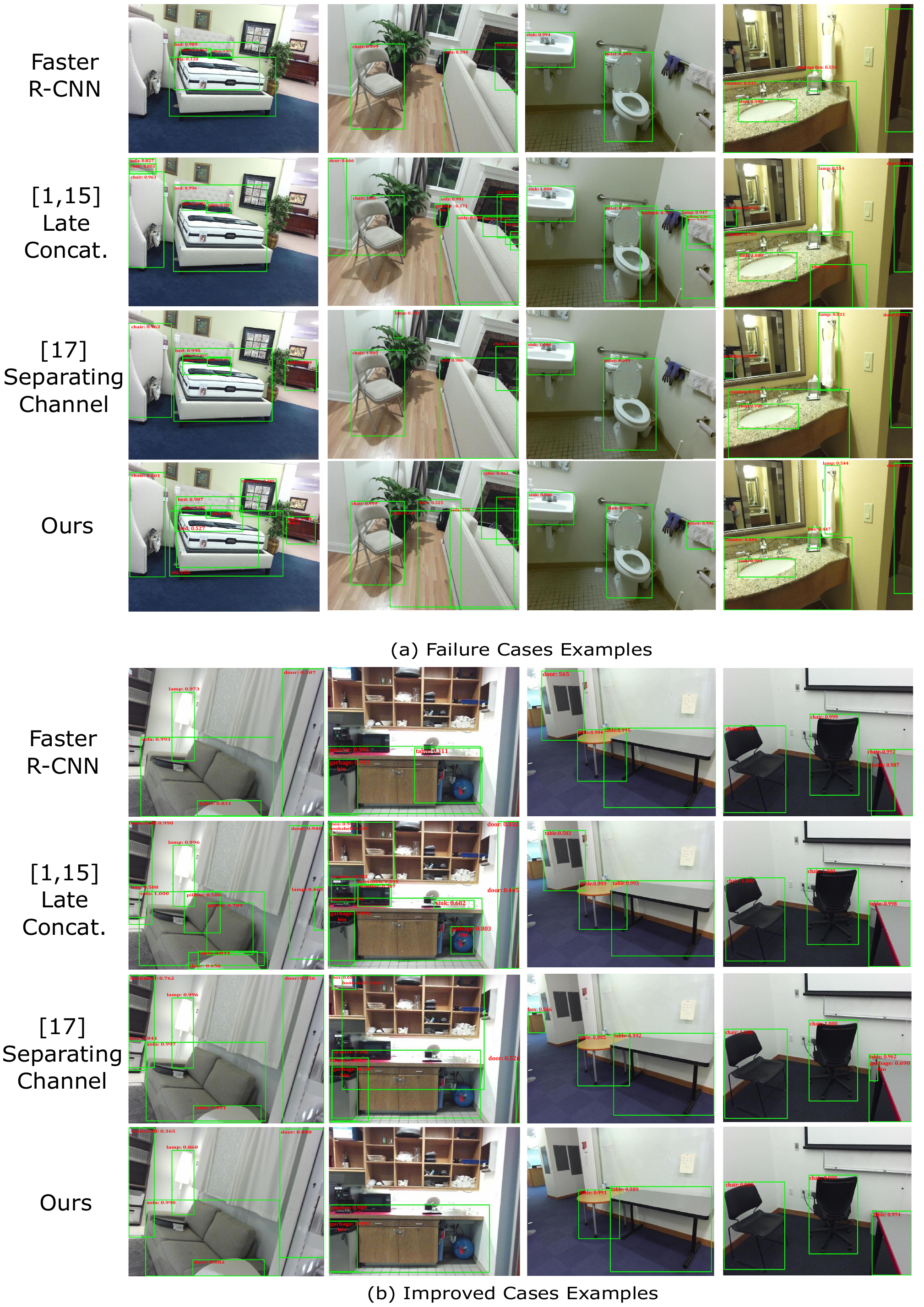}
    \caption{Examples of visual results for failure, part (a), and improved, part(b), cases. Images are taken from SUN RGB-D test split. The first rows in part (a) and (b) results belong to Faster R-CNN. The second results are belongs to Late Concatenation, shown in Figure \ref{fig:All_Architecture} part (b)-3 proposed in \cite{cao2017exploiting,gupta2014learning}. The idea of processing each channel of HDHA encoding separately shown in Figure \ref{fig:PrevArchitecture} which is proposed in \cite{hou2018object}, generates third rows results. Our proposed architecture shown in \ref{fig:All_Architecture}, part (b)-1, generates the last rows results. HDHA encoding and ResNet-101 backbone are used to generate these outputs. }
    \label{fig:ImprovedEx}
\end{figure*}

\begin{figure*}[hbt!]
   \centering
    \includegraphics[width=0.8\textwidth]{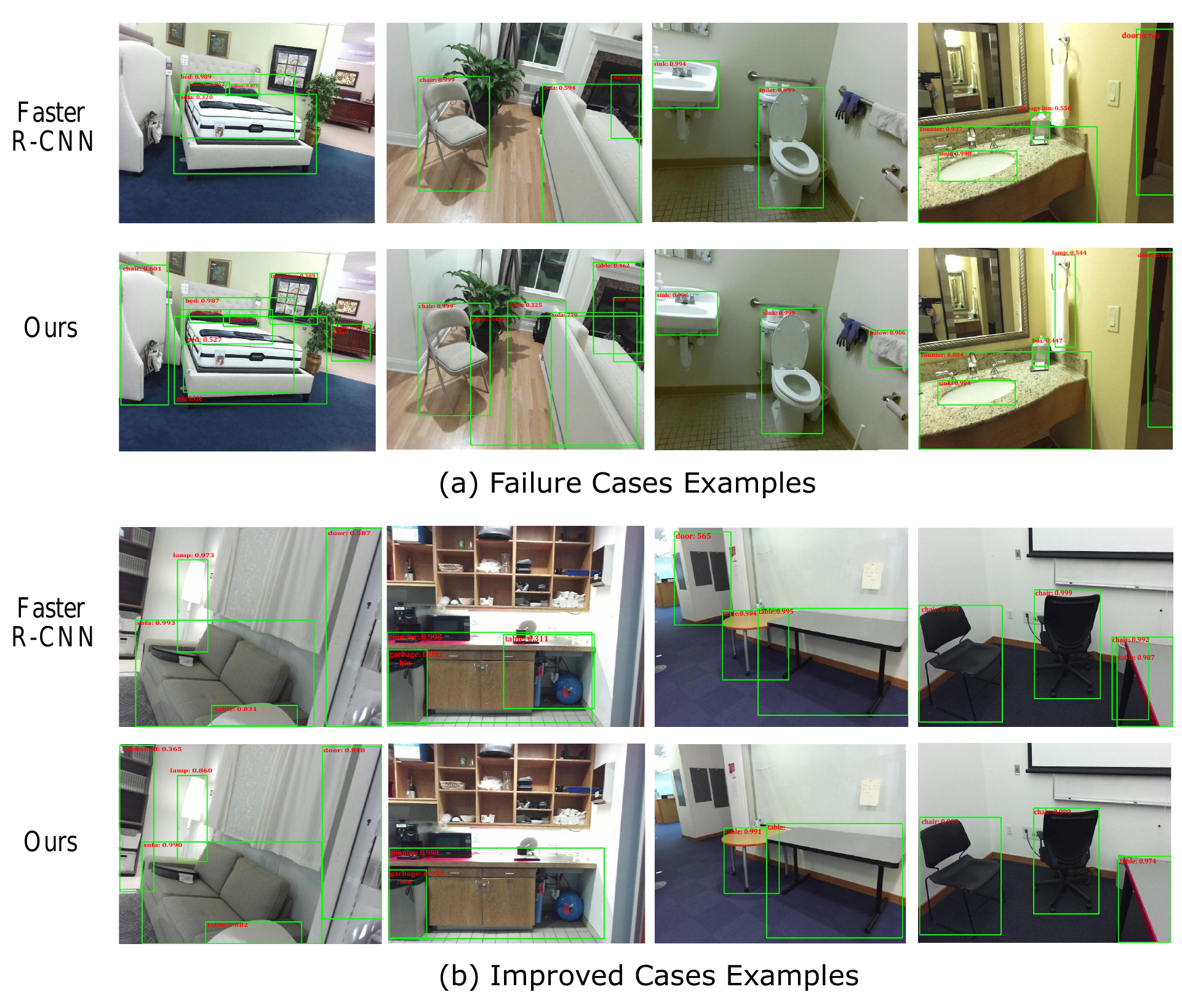}
    
    \caption{Examples of visual results for failure, part (a), and improved, part(b), cases. Images are taken from SUN RGB-D test split. The first rows in part (a) and (b) results belong to Faster R-CNN and the last rows show our results. In the first row of part (a), our model detected non-existing objects in the scene such as television, night-stand and chair. Like this results, it also detected chair,tables, and door which are false-positive detection in the second row result. In the third row result, we classified a toilet as a sink and found non-existing pillow. Finally, it detected non-existing lamp in the fourth row result. For improved cases shown in part (b), Our model detected bookshelf which was not detected by the Faster R-CNN in the first column. Also, it removes a table, door and chair detections in the second, third and fourth rows, respectively.  }
    \label{fig:ImprovecdExx}
\end{figure*}

\section{DISCUSSION AND CONCLUSION}
\label{sect:conclusion}
In this study, we revisited RGB-D object detection problem from different angles. With systematic experiments, we clarified the following questions: (i) Does depth improve object detection? (ii) For what kind of objects depth is helpful? (iii) How and when should the depth information be integrated into the object detection pipeline?

\subsection{Does depth improve object detection?}
There is not any simple answer to this question, because there are numerous factors that affect the contribution of the depth map to object detection performance. We investigate these effects from following aspects: (i) backbone network, (ii) depth processing, (iii) depth encodings and (iv) concatenation points in this work.

\subsubsection{Backbone Network}
According to our experimental results, shown in Tables \ref{table:SUNRGB-D},  \ref{table:VOC2007Indoor} and \ref{table:VOC2007}, VGG-16 has lower object detection performance than ResNet-101 for Faster R-CNN.

For the SUN RGB-D dataset, depth information improves Faster R-CNN performance without being affected by the backbone network choice and the depth encoding type. 

Also, previous work proposed by Cao et al. \cite{cao2017exploiting} could improve Fast R-CNN performance by using VGG-16 backbone and estimated depth maps for Pascal VOC dataset. Unlike this result, we could not improve the Faster R-CNN performance with any backbone choice and the depth encoding type and estimated depth maps, although we conduct detailed experiments, presented in Sections S3 and S4, to estimate depth map.

Based on both ours and Cao et al.'s \cite{cao2017exploiting} experiments, we can conclude that, as the performance of the object detection network increases, the benefits of depth information decrease. 

\subsubsection{Depth Encodings}
First, the major grouping in depth is whether it is  ground-truth or  estimated. Then, each group has three different encodings, which are gray-scale, HDHA, and jet color space. As expected, ground-truth depth  always improves Faster R-CNN performance. Moreover, all depth representations have  nearly the same performance, although HDHA encoding contains richer information than the other representations. Unlike ground-truth depth, estimated depth does not always improve Faster R-CNN performance. It is heavily affected by the dataset. While it improves Faster R-CNN performance on the SUN RGB-D dataset, it does not do so on the PASCAL VOC and COCO datasets regardless of  backbones and encodings. This result on PASCAL and COCO needs a closer inspection which we do in the following.

\begin{figure*}[h]
  \includegraphics[width=\linewidth]{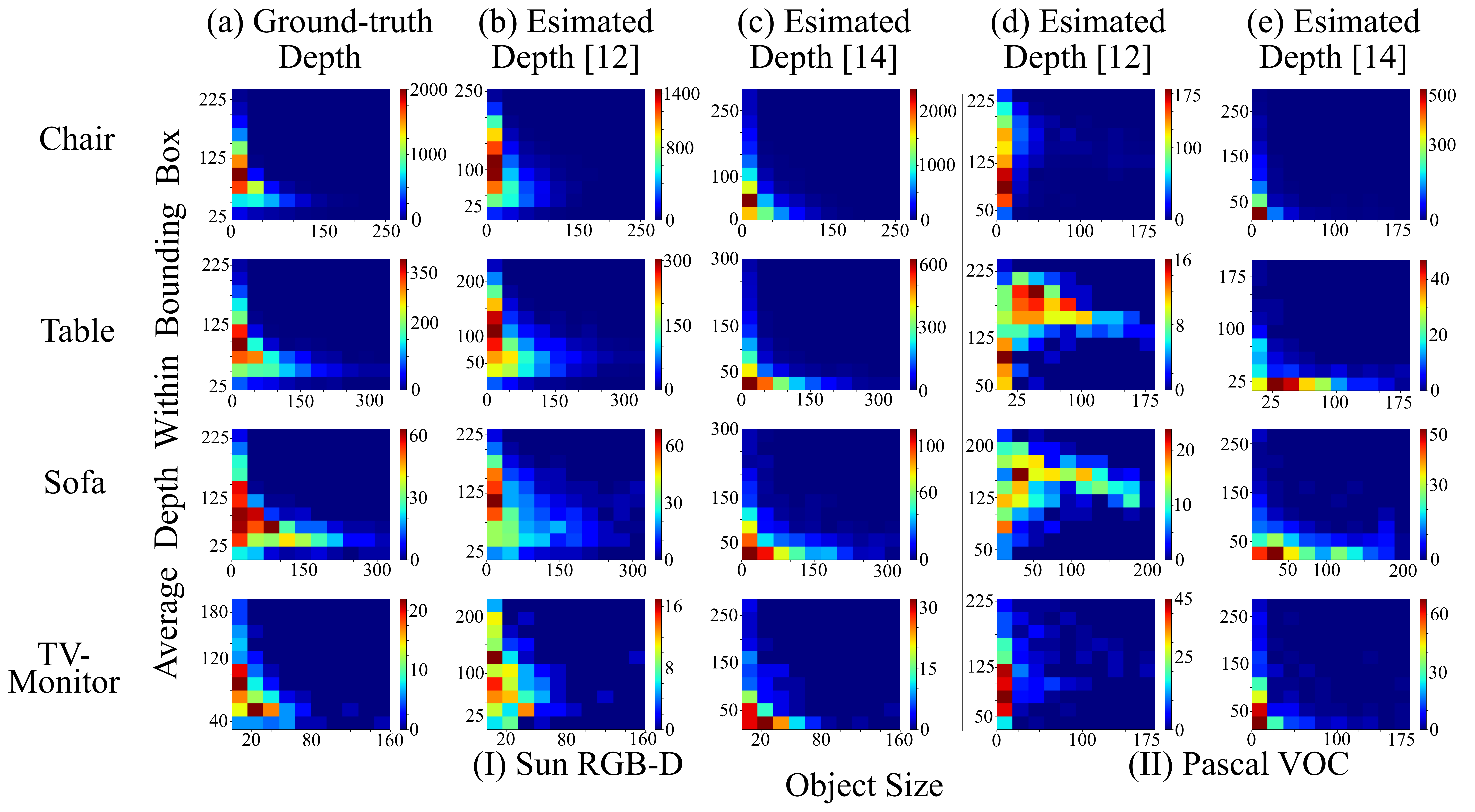}
  \caption{
  Average depth vs Object size for common classes in SUN RGB-D and PASCAL VOC. 
  Estimated depth improves object detection performance on the SUN RGB-D dataset (Table \ref{table:SUNRGB-D}). This is supported by the fact  that the heatmap patterns in columns (b) and (c) are similar to the patterns of the ground-truth column (a). On the other hand, estimated depth does not improve object detection performance on PASCAL VOC (Table \ref{table:VOC2007}), which can be explained by the dissimilar patterns of columns (d) and (e) to that of (a). For the sake of visibility, each $x$ and $y$ axes have a local scale.
  }
  \label{fig:MeanHeatmap}
\end{figure*}

PASCAL VOC object detection results using different depth estimation networks (DEN) are given in Table S1. None of the DENs improve performance; on the contrary, they degrade the performance. We argue that the reason behind this is the   poor quality of depth estimation. Directly measuring this quality  using depth metrics is not possible due to the absence of ground-truth depth data. Therefore, we turn to SUN RGB-D, which has ground-truth depth data, and base our analysis on the object classes that are common in both SUN RGB-D and PASCAL VOC. These classes are  chair, table, sofa, and television. For the instances pertaining to these classes, in Figure \ref{fig:MeanHeatmap}, we plot the average depth  versus size (bounding box area) shown for two state-of-the-art DENs \cite{hu2019revisiting, miangoleh2021boosting}. In general, while the estimated depth plots for SUN RGB-D (columns b, c) are very similar to the ground-truth (column a), PASCAL plots (columns d,e) are not. For the ground-truth depth plots (part (a)), an inverse correlation pattern can be observed: as the object size increases, there are more objects that have low average depth. This pattern can generally be observed in other columns, except the column (d) which is the estimated depth  \cite{hu2019revisiting} for PASCAL VOC. The similarity between column (a) (ground-truth depth) and column (b) (estimated depth \cite{hu2019revisiting} for SUN RGB-D) is especially striking. Although the aforementioned pattern can be observed in column (e), which is estimated depth \cite{miangoleh2021boosting} for PASCAL VOC, this depth estimation has a scaling problem as shown in column (e) of Figure \ref{fig:MeanHeatmap}. While pixel values of these estimated depths \cite{miangoleh2021boosting} have a smaller range for chair and TV-Monitor classes in Pascal VOC dataset, i.e. most of average depth within bounding box is between 0 and 50 for these classes, it has closer results to ground-truth data in SUN RGB-D datasets shown in column (c) of Figure \ref{fig:MeanHeatmap}. That is why, we improve the Faster R-CNN performance for only SUN RGB-D dataset using estimated depth. Hence, the effect of depth to object detection performance is heavily affected by depth map quality.

\subsection{How does using  depth affect object detection?}
To investigate this effect, we analyse the confusion matrix of our Faster R-CNN on SUN RGB-D in Table S2. Here we observe that  ground-truth depth information helps to detect objects which cannot be detected using RGB features. Therefore, the main contribution of ground-truth depth encodings is decreasing number of false negatives as shown in the last column of Table S2. This contribution is present for all classes in the SUN RGB-D dataset.

The second effect of ground-truth depth is on confusion between classes. For example, SUN RGB-D dataset contains classes that have similar RGB appearances with other classes such TV-monitor and desk-table. Depth information reduces the confusion between these classes in favor of the class with more examples. Desk objects are detected more as tables, while monitor objects are predicted more often as televisions. 

However, this confusion does not prevent depth from improving the performance Faster R-CNN for all classes.

On the other hand, the estimated depth increases false negative cases for most classes in PASCAL VOC and SUN RGB-D datasets as shown in Tables S3 and S4 due to poor depth map quality.

\subsection{How and when should the depth information be integrated into the object detection pipeline?}

Our method, Early Concatenation, has better performance than other concatenation types for all depth encoding types. Also, Early Concatenation has the advantage of using fewer parameters than others.

However, when the ground-truth depth is used, processing each channel of HDHA encoding separately gives a better performance than our method.
If model complexity is not an important factor and ground-truth HDHA data are available, processing each channel of HDHA encodings separately should be chosen. For the rest of the cases, our method is more suitable.

\bibliographystyle{elsarticle-num}
\bibliography{main}

\end{document}


\newcommand{\CStart}[0]{\begin{tabular}[c]{@{}c@{}}}
\newcommand{\CEnd}[0]{\end{tabular}}
\newcommand\RotText[1]{\fontsize{9}{9}\selectfont \rotatebox[origin=c]{90}{\parbox{2.6cm}{\centering#1}}}

\newcolumntype{C}{@{\hspace{1pt}}c@{\hspace{1pt}}}

\newcommand{\BC}[1]{\textcolor{teal}{[BC: #1]}}
\newcommand{\EA}[1]{\textcolor{blue}{[EA: #1]}}
\newcommand{\SK}[1]{\textcolor{red}{[SK: #1]}}
\renewcommand{\thesection}{S.\arabic{section}}
\renewcommand{\thetable}{S.\arabic{table}}
\renewcommand{\thefigure}{S.\arabic{figure}}

\begin{frontmatter}

\title{Supplementary Material}

\end{frontmatter}

\section{Training and Implementation Details}

During experiments, Faster R-CNN implementation from \cite{jjfaster2rcnn} was used. The pre-trained models on ImageNet \cite{russakovsky2015imagenet} from  \cite{jjfaster2rcnn} were used for backbones VGG-16  and ResNet-101. For VGG-16, the layers before the \textit{conv3} were fixed. For ResNet-101, the first block and batch normalization layers were frozen.

While RGB images are normalized with the mean of ImageNet dataset, other inputs (depth encodings) are normalized with their mean over the training set.

The short side of all images was scaled to 600 in training and testing stages. During training, a horizontal flip operation was used for data augmentation. After applying non-maximum suppression (NMS), the top 2000 scoring boxes were selected in training and the top 300 scoring boxes were selected in the testing stage. 

For training the networks in all experiments, Stochastic Gradient Descent with a momentum rate of $0.9$, a learning rate of $0.001$, and a batch size of 1 were used. After every $5$ epochs, the learning rate was decayed by a factor of $0.1$. Using early stopping, the baseline Faster R-CNN was trained for $6$ epochs on Pascal VOC 2007 and SUN-RGBD datasets whereas our RGB-D object detection network was trained for $7$ epochs. For training and testing the networks, a machine with Tesla v100 was used.

\section{Concatenation Details}
\label{subsec:ConcatDetails}
\subsection{Raw Depth}
To integrate raw depth images to the object detection pipeline, we should match the size of depth images to related RGB features size which depends on the concatenation type. Therefore, each concatenation type requires different resizing and convolution operations. 

In Early Concatenation, the size of RGB features is 1024$\times$w$\times$h, where \textit{w} and \textit{h} depend on the size of RGB images. We resize each depth image to corresponding \textit{w} and \textit{h}. Then, we use one convolution layer that has 1024 filters and 1$\times$1 kernel to match channel count. After these operations, we concatenate raw depth map and RGB features along the channel axis, i.e. the size of fused features is 2048 $\times$w$\times$h. Finally, we decrease the number of fused features channels to 1024 using another 1$\times$1 convolution layer. The rest of the object detection pipeline is the same as Faster R-CNN.

Unlike Early Concatenation, the size of RGB features which is 1024$\times$7$\times$7 is the same for all RGB images in Middle-Late Concatenation because the RoI Align module produces fixed size RGB features. Therefore, we resize each depth image to 7$\times$7. To match channel count, we use the convolution layer which has 1024 filters with size 1$\times$1. Also, we repeat depth images $n$ times, where $n$ is the number of predicted objects. After these operations, we concatenate depth images and RGB features in the channel axis, i.e. size of the fused features is $n \times$2048$\times$7$\times$7. Finally, we decrease the number of fused channels like in Early Concatenation. The rest of the object detection pipeline is the same as Faster R-CNN.

In Late Concatenation, RGB features have n$\times$4096, where n is number of predicted objects. Therefore, we first resize each depth image to 64$\times$64, then flatten it. To integrate raw depth images to RGB features, we just repeat depth image n times. Then, we concatenate them in batch axes. Bounding box prediction and classification tasks are performed on these fused features.

\subsection{Processed Depth}
Unlike the raw depth case, processed depth does not require any additional operation to match size with RGB features because we use separate backbone networks to extract depth features and these features already have the same size with RGB features for all concatenation types. Therefore, we just concatenate depth features and RGB features along the channel dimension.

\section{Depth Estimation Network Experiments}
\label{section:DepEstExp}

Widely used Pascal VOC and MS-COCO datasets do not provide ground-truth depth. Therefore, if one wants to use depth information for object detection, it should be estimated from single RGB images. For this purpose, we evaluated the performances of the following state-of-the-art depth estimation networks:  \cite{hu2019revisiting, li2018megadepth, miangoleh2021boosting, watson2019self, lasinger2019towards}.

\begin{figure}
\centering
{
\scriptsize
\renewcommand{\arraystretch}{0.3}
\newcommand{\SKALA}[0]{0.1}
\begin{tabular}{ m{1.5cm}  m{1.2cm} m{1.2cm} m{1.2cm} m{1.2cm}}
\CStart{}Input\CEnd{}         & 
    \includegraphics[scale=\SKALA]{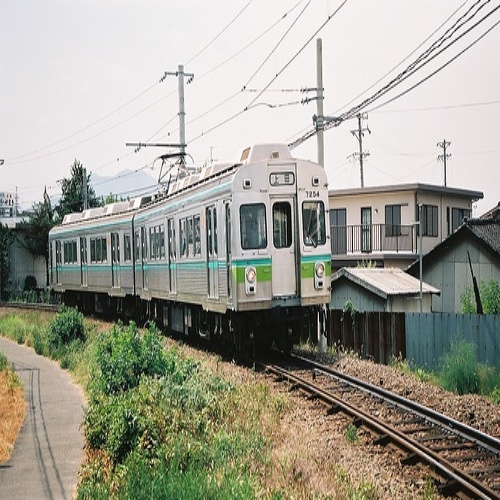} &
    \includegraphics[scale=\SKALA]{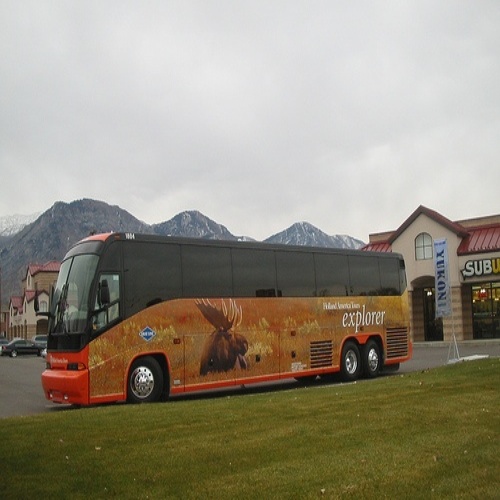} & 
    \includegraphics[scale=\SKALA]{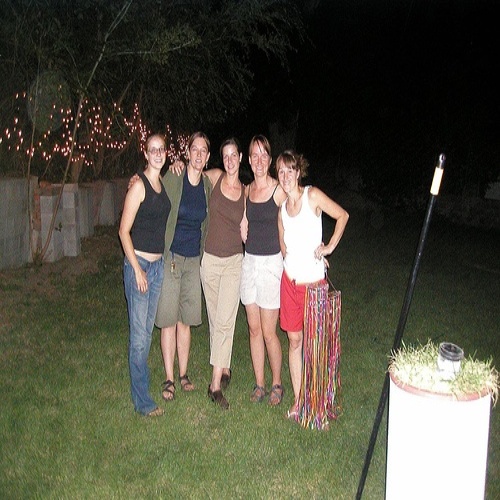} & 
    \includegraphics[scale=\SKALA]{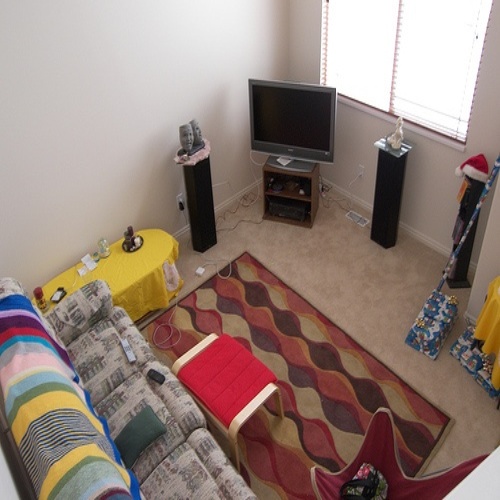} \\\\
\CStart{} MonoDepth \cite{watson2019self} \CEnd{}         &
    \includegraphics[scale=\SKALA]{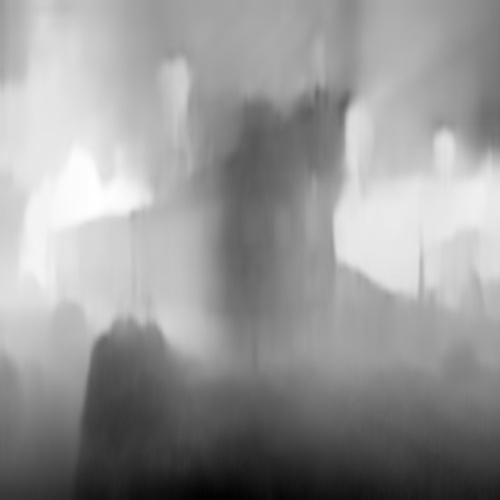} &
    \includegraphics[scale=\SKALA]{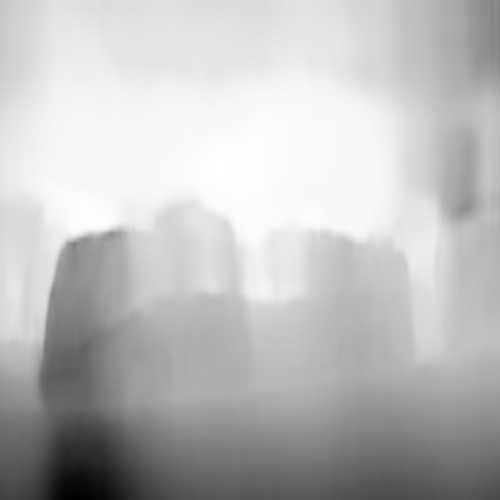} & 
    \includegraphics[scale=\SKALA]{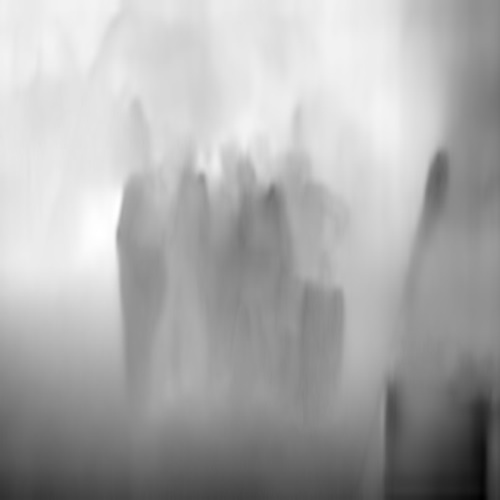} &
    \includegraphics[scale=\SKALA]{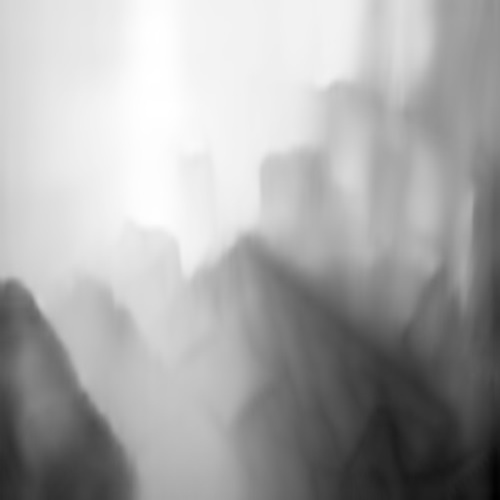} \\\\
\CStart{} MegaDepth \cite{li2018megadepth} \CEnd{}         & 
    \includegraphics[scale=\SKALA]{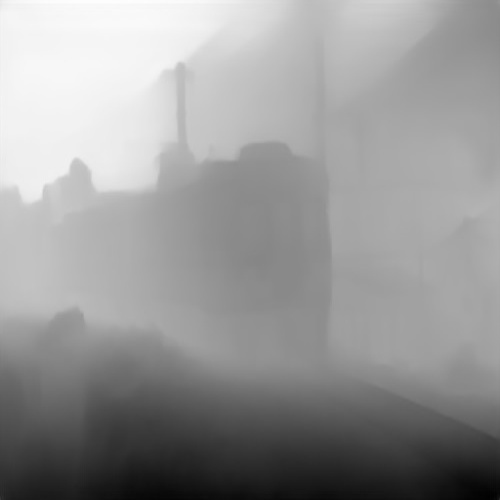} &
    \includegraphics[scale=\SKALA]{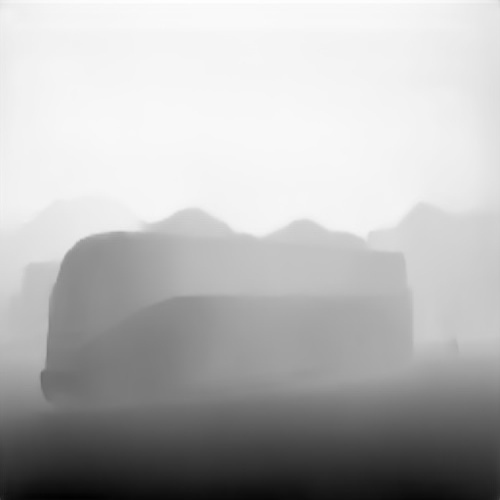} & 
    \includegraphics[scale=\SKALA]{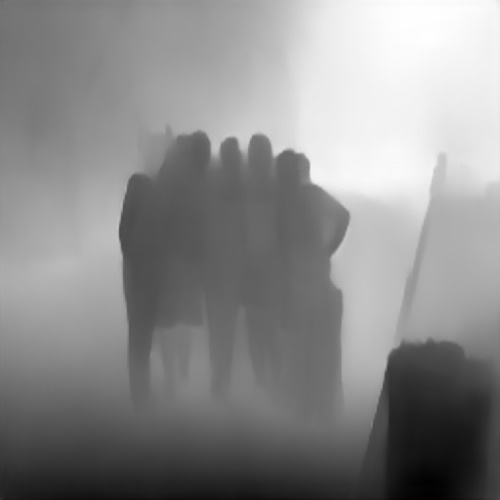} &
    \includegraphics[scale=\SKALA]{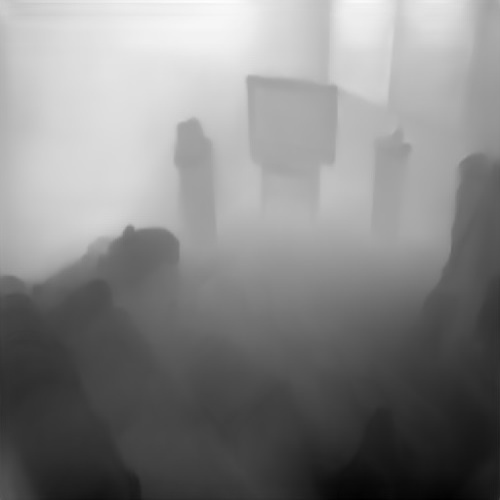} \\\\
\CStart{}Boundary \\ Depth \cite{hu2019revisiting} \CEnd{}         & 
    \includegraphics[scale=\SKALA]{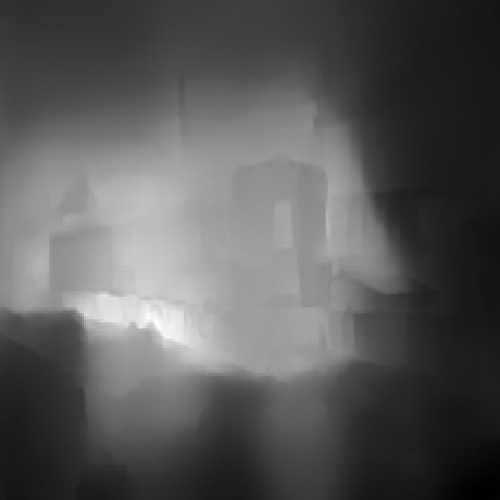} &
    \includegraphics[scale=\SKALA]{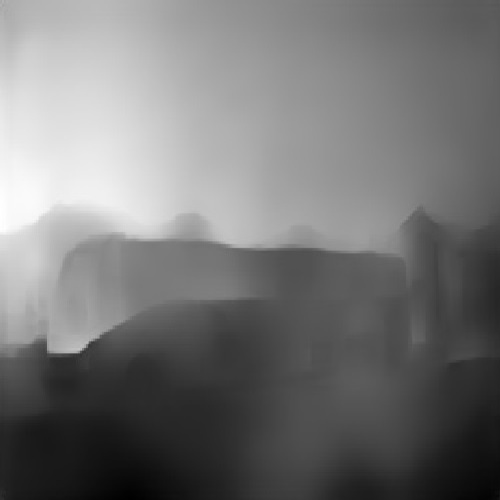} & \includegraphics[scale=\SKALA]{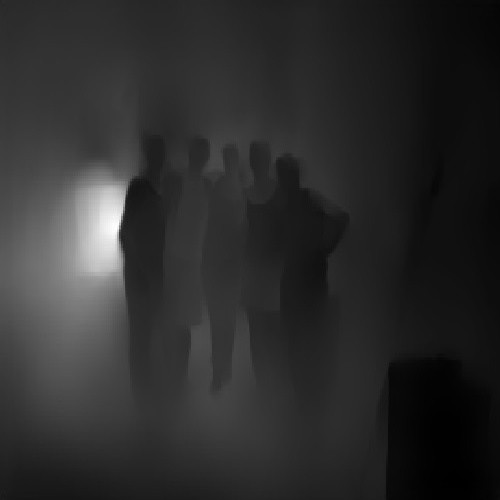} &
    \includegraphics[scale=\SKALA]{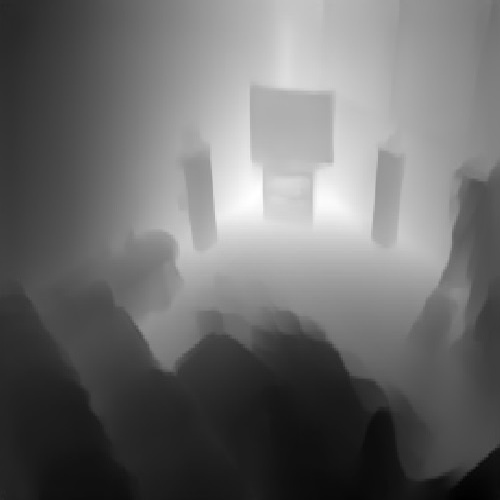} \\\\
\CStart{}Midas \cite{lasinger2019towards}\CEnd{}         & 
    \includegraphics[scale=\SKALA]{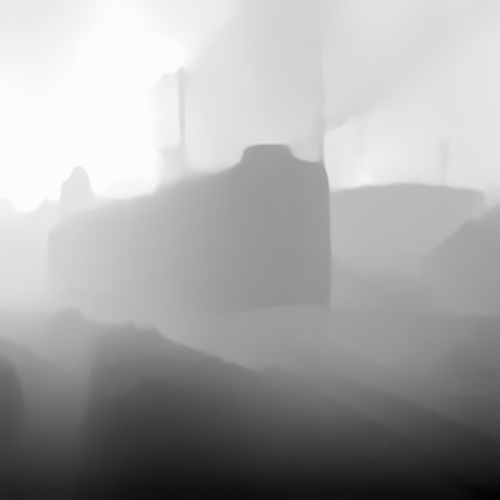} &
    \includegraphics[scale=\SKALA]{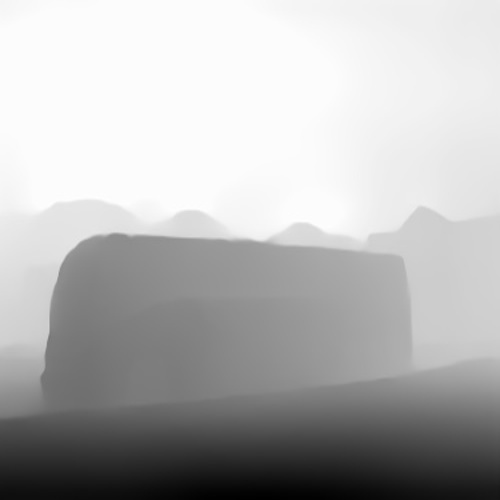} & 
    \includegraphics[scale=\SKALA]{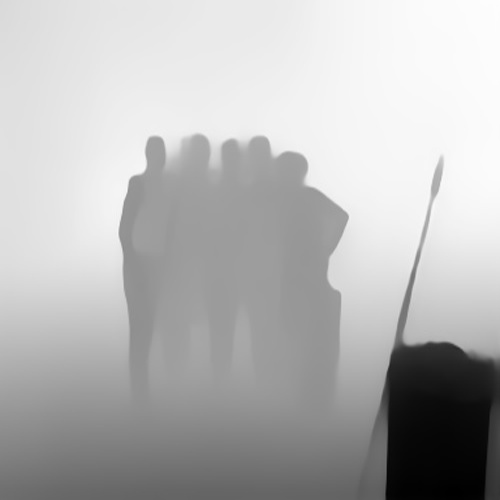} &
    \includegraphics[scale=\SKALA]{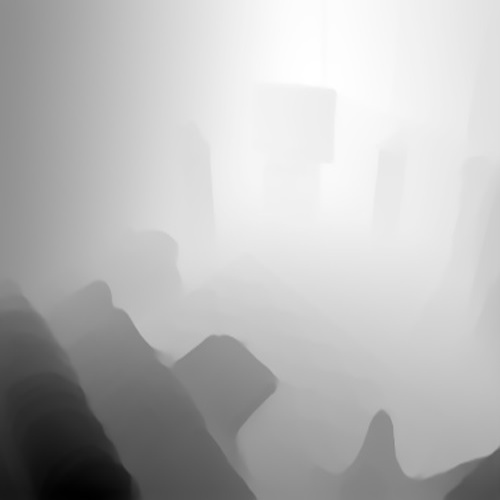} \\\\
\CStart{}Boosting \cite{miangoleh2021boosting}\CEnd{}         & 
    \includegraphics[scale=\SKALA]{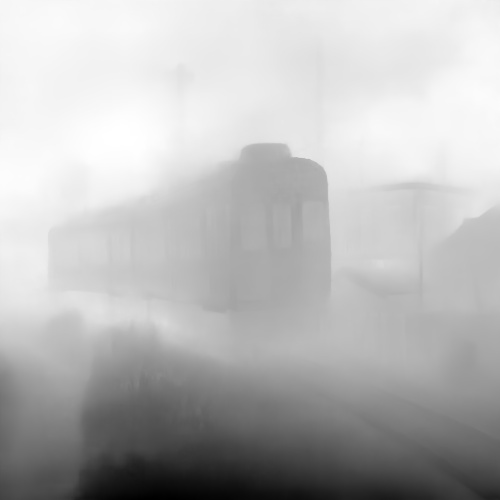} &
    \includegraphics[scale=\SKALA]{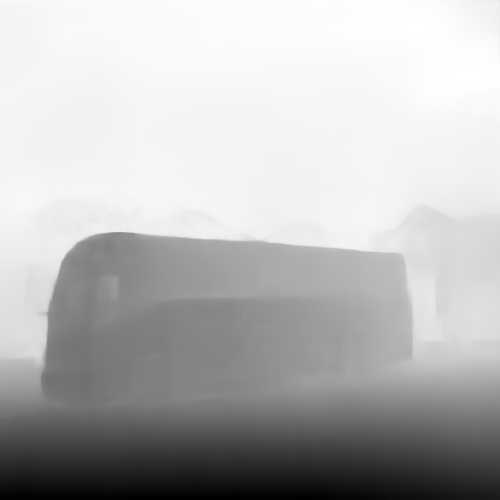} & 
    \includegraphics[scale=\SKALA]{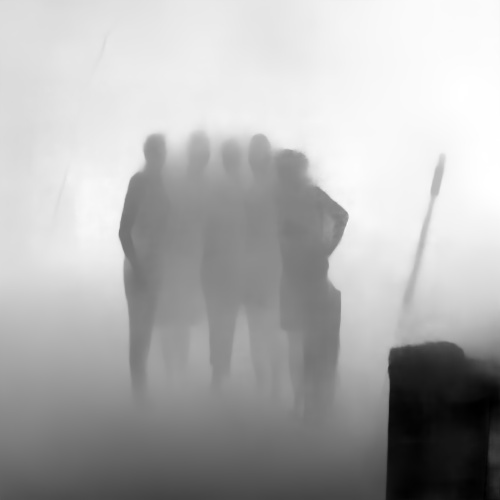} &
    \includegraphics[scale=\SKALA]{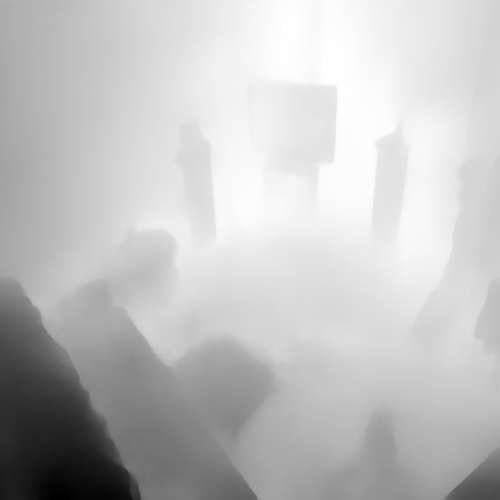} \\\\
\end{tabular}
}
\caption{Depth estimation network outputs for images example from Pascal VOC 2007}
\label{fig:DepEst2}
\end{figure}

\begin{figure}
\centering
{
\scriptsize
\renewcommand{\arraystretch}{0.3}
\newcommand{\SKALA}[0]{0.1}
\begin{tabular}{ m{1.5cm}  m{1.2cm} m{1.2cm} m{1.2cm} m{1.2cm}}
\CStart{}Input\CEnd{}         & 
    \includegraphics[scale=\SKALA]{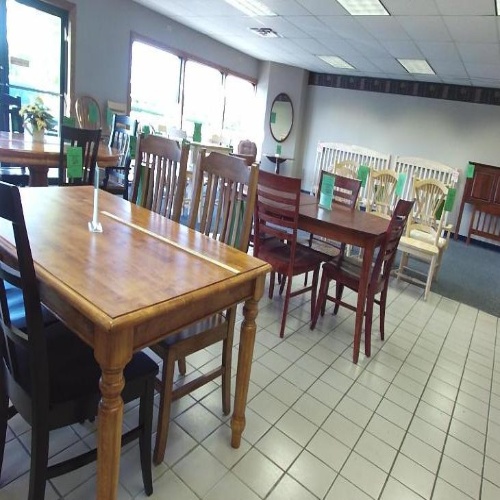} &
    \includegraphics[scale=\SKALA]{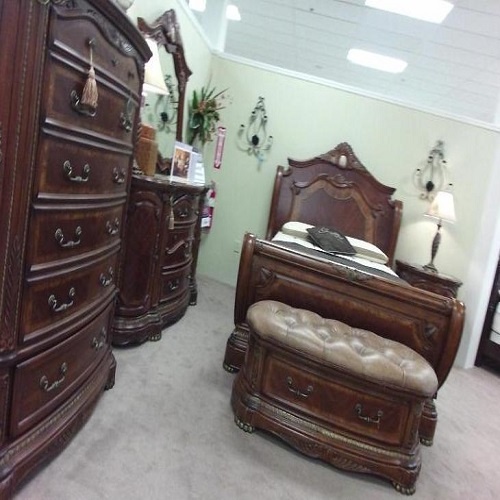} & 
    \includegraphics[scale=\SKALA]{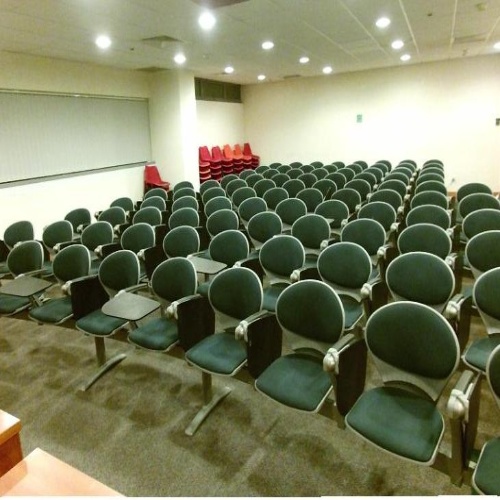} & 
    \includegraphics[scale=\SKALA]{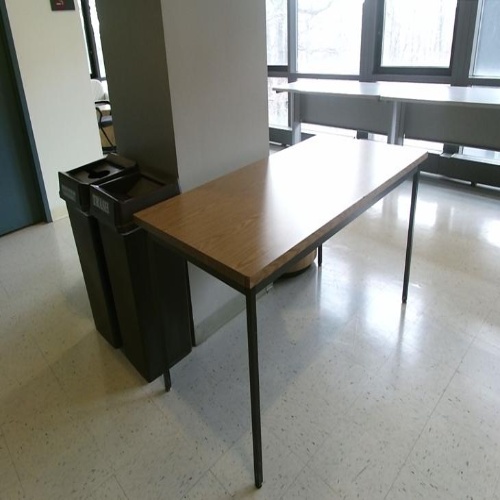} \\\\
\CStart{}Ground-Truth \\ Depth\CEnd{}         & 
    \includegraphics[scale=\SKALA]{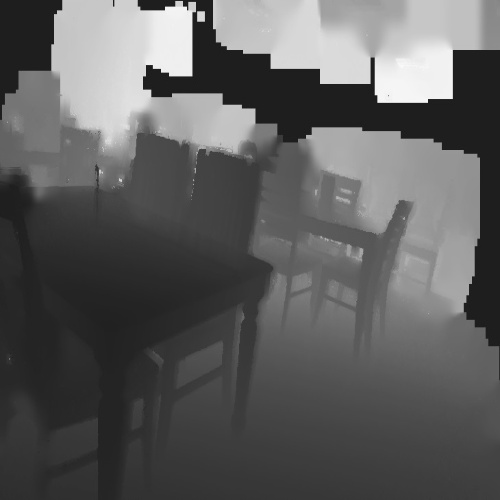} &
    \includegraphics[scale=\SKALA]{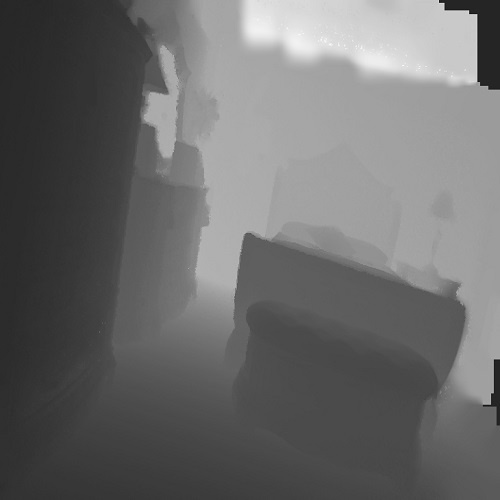} & 
    \includegraphics[scale=\SKALA]{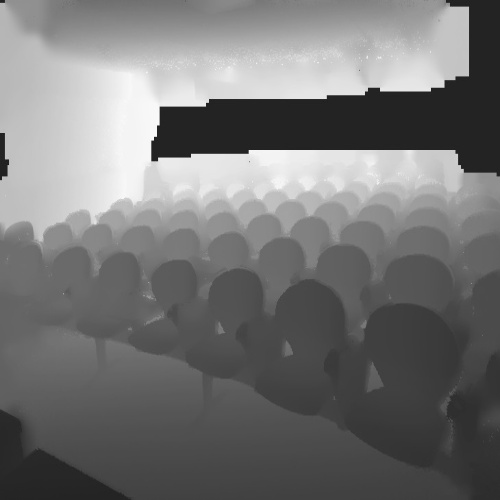} & 
    \includegraphics[scale=\SKALA]{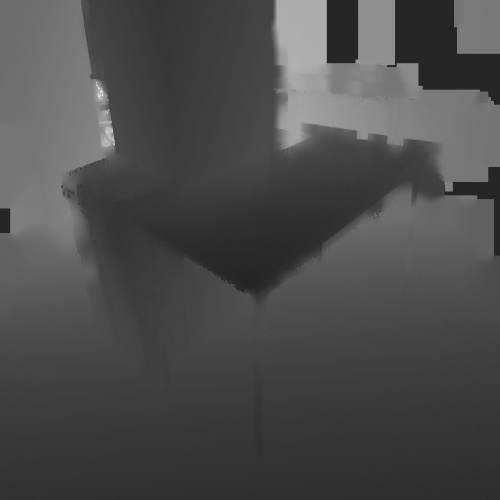} \\\\
\CStart{} MonoDepth \cite{watson2019self} \CEnd{}         &
    \includegraphics[scale=\SKALA]{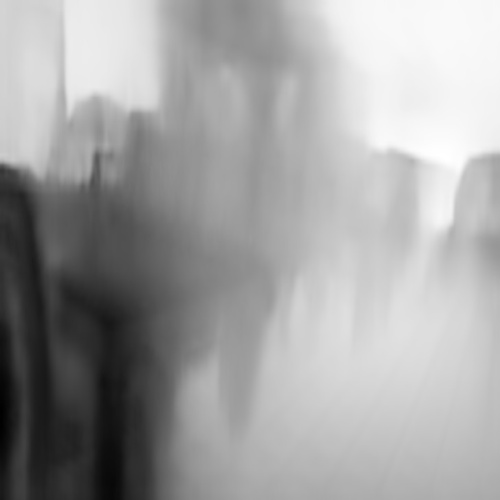} &
    \includegraphics[scale=\SKALA]{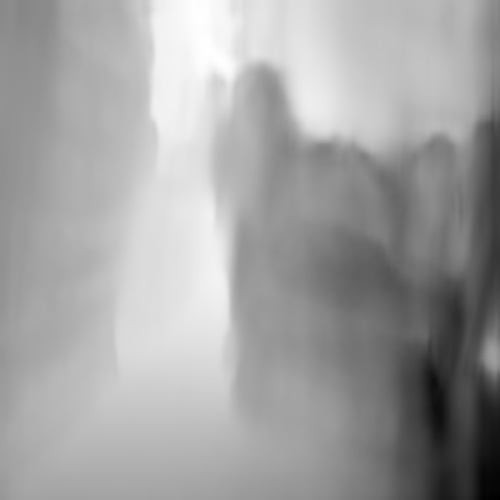} & 
    \includegraphics[scale=\SKALA]{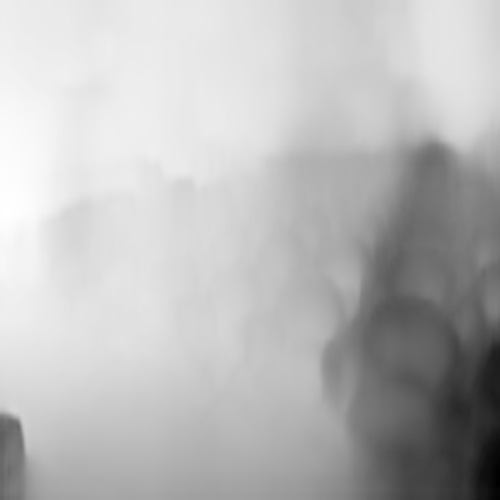} &
    \includegraphics[scale=\SKALA]{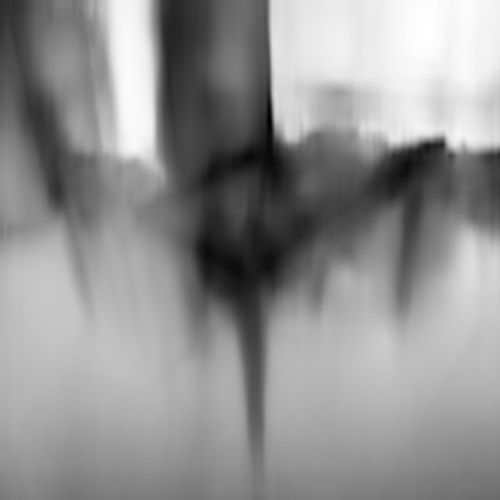} \\\\
\CStart{} MegaDepth \cite{li2018megadepth} \CEnd{}         & 
    \includegraphics[scale=\SKALA]{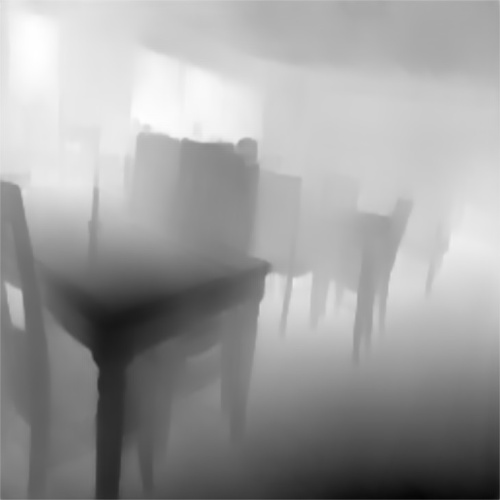} &
    \includegraphics[scale=\SKALA]{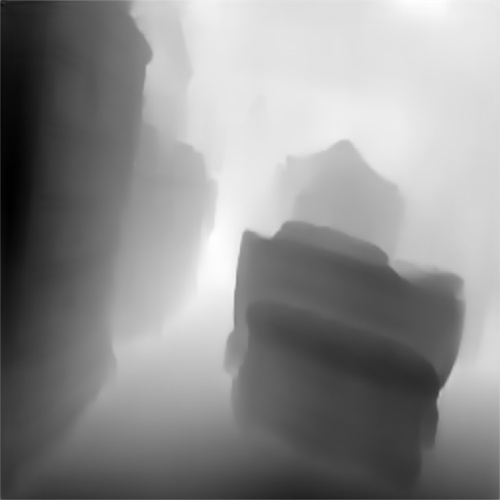} & 
    \includegraphics[scale=\SKALA]{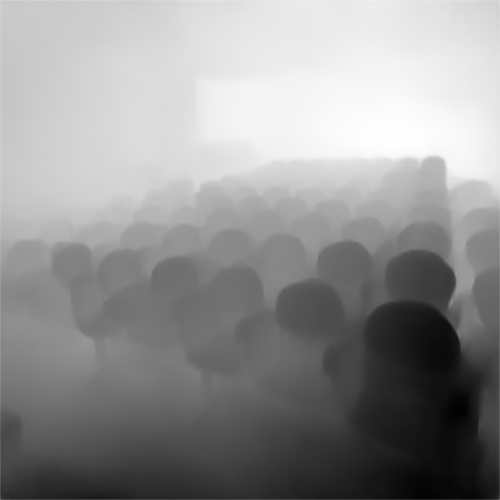} &
    \includegraphics[scale=\SKALA]{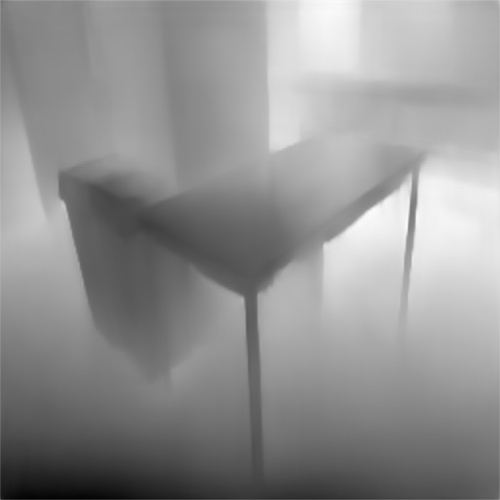} \\\\
\CStart{}Boundary \\ Depth \cite{hu2019revisiting} \CEnd{}         & 
    \includegraphics[scale=\SKALA]{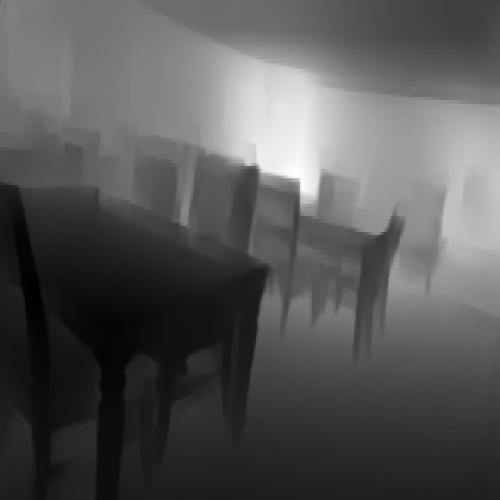} &
    \includegraphics[scale=\SKALA]{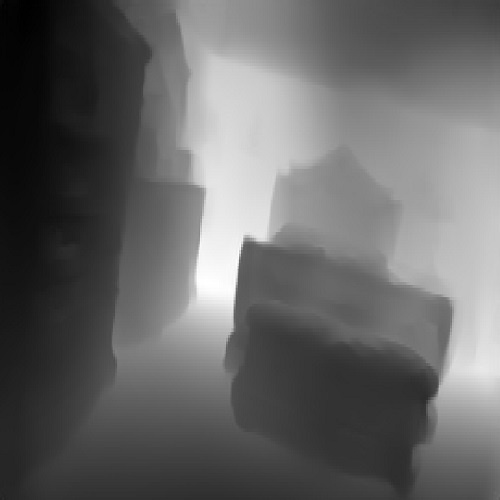} & 
    \includegraphics[scale=\SKALA]{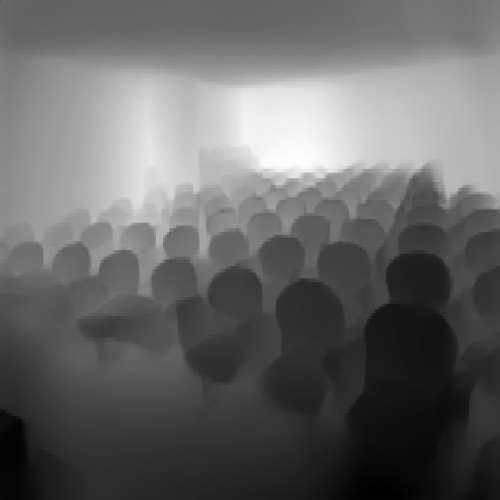} &
    \includegraphics[scale=\SKALA]{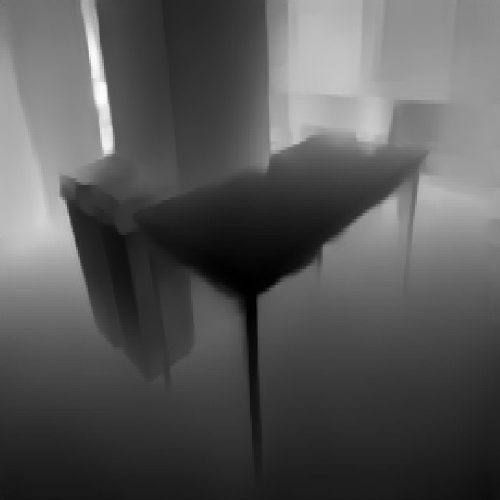} \\\\
\CStart{}Midas \cite{lasinger2019towards}\CEnd{}         & 
    \includegraphics[scale=\SKALA]{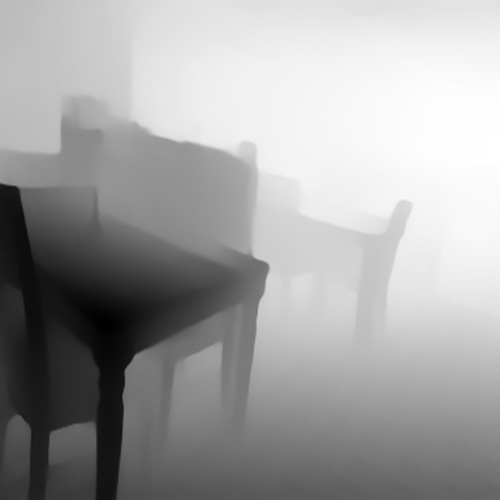} &
    \includegraphics[scale=\SKALA]{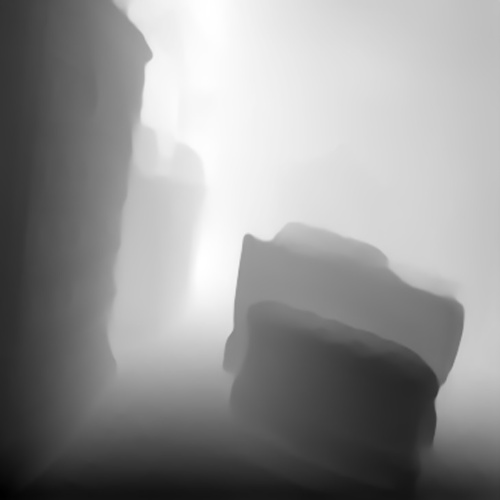} & 
    \includegraphics[scale=\SKALA]{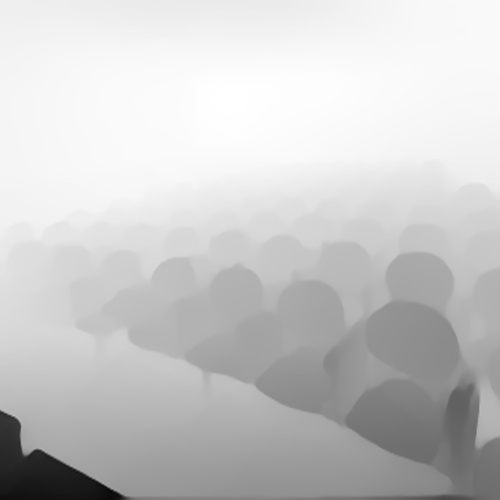} &
    \includegraphics[scale=\SKALA]{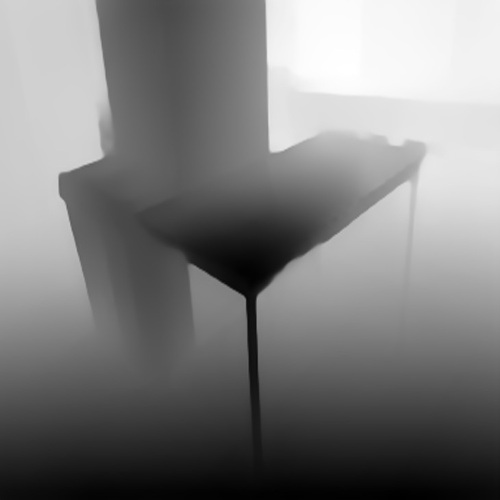} \\\\
\CStart{}Boosting \cite{miangoleh2021boosting}\CEnd{}         & 
    \includegraphics[scale=\SKALA]{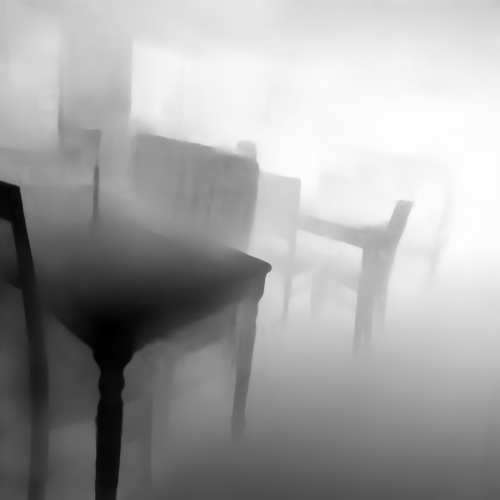} &
    \includegraphics[scale=\SKALA]{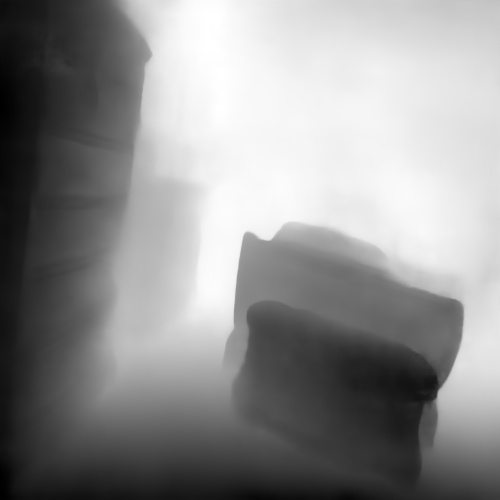} & 
    \includegraphics[scale=\SKALA]{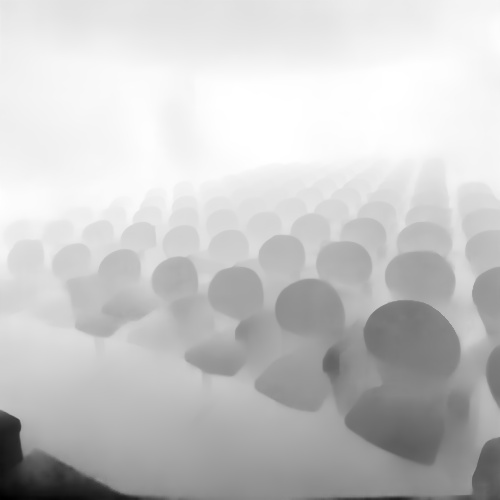} &
    \includegraphics[scale=\SKALA]{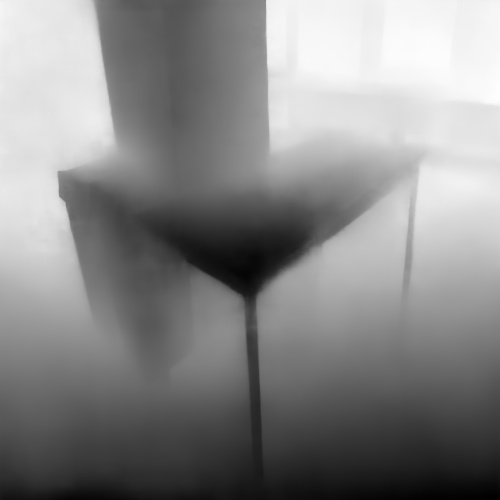} \\\\
\end{tabular}
}
\caption{Depth estimation network outputs for images example from SUN RGB-D test set.}
\label{fig:DepEst}
\end{figure}

Figures \ref{fig:DepEst2} and \ref{fig:DepEst}  show sample depth estimations from these networks. For all networks, official pre-trained models are used.

Qualitatively, the method proposed by Miangoleh et al. \cite{miangoleh2021boosting} produces the best results as seen in the last row of  Figures \ref{fig:DepEst2} and \ref{fig:DepEst}. However, their output is a relative depth map and this leads to scaling problem, when we tried to create absolute depths from its outputs. Therefore, most of object detection experiments were done with the method proposed by Hue et al. \cite{hu2019revisiting}, since it produces absolute depth map and more sharper objects in the estimated depth map than the other methods. 

\begin{table*}[h]
\caption{Object Detection Results on Pascal VOC dataset. The first column shows results of Faster R-CNN. Remaining rows shows estimated grayscale depth maps with RGB images results on Pascal VOC dataset.
} 
\label{table:PascalVOC_DEN}
\centering
\begin{adjustbox}{width=1\textwidth}
\small
\begin{tabular}{c|cccccccccccccccccccc|c}
\hline
  & \rotatebox{90}{a.plane} & \rotatebox{90}{bicycle} & \rotatebox{90}{bird} & \rotatebox{90}{boat} & \rotatebox{90}{bottle} & \rotatebox{90}{bus} & \rotatebox{90}{car} & \rotatebox{90}{cat} & \rotatebox{90}{chair} & \rotatebox{90}{cow} & \rotatebox{90}{d.table} & \rotatebox{90}{dog} & \rotatebox{90}{horse} & \rotatebox{90}{m.bike} & \rotatebox{90}{person} & \rotatebox{90}{p.plant} & \rotatebox{90}{sheep} & \rotatebox{90}{sofa} & \rotatebox{90}{train} & \rotatebox{90}{tv/mon.} & \rotatebox{90}{mAP}\hfil \\
 \hline
\hline
Base  & 76.3 & 81.8 & 77.0 & 66.2 & 60.6 & 80.2 & 85.5 & 86.5 & 55.6 & 83.3 & 65.6 & 85.8 & 85.3 & 77.9 & 78.6 & 48.2 & 75.9 & 74.2 & 77.9 & 76.2 & 74.9
\\
\hline

MonoDepth2 \cite{watson2019self}  & 77.6 & 81.6 & 73.9 & 60.6 & 57.0 & 83.6 & 86.7 & 86.1 & 55.0 & 78.5 & 65.9 & 85.5 & 83.2 & 78.0 & 79.0 & 48.8 & 77.0 & 72.8 & 80.5 & 73.6 & 74.3  \\
\hline
MegaDepth \cite{li2018megadepth} & 76.4 & 81.3 & 77.6 & 65.7 & 60.0 & 83.1 & 85.1 & 85.0 & 54.0 & 79.1 & 68.0 & 86.0 & 83.2 & 75.6 & 78.8 & 47.6 & 75.6 & 75.2 & 78.7 & 73.1 & 74.4    \\

\hline
BoundaryDepth \cite{hu2019revisiting}  & 78.3 & 79.8 & 77.0 & 63.6 & 59.5 & 82.9 & 85.9 & 85.8 & 55.3 & 76.4 & 66.0 & 84.4 & 83.8 & 77.7 & 78.7 & 47.2 & 75.5 & 73.5 & 79.8 & 73.5 & 74.2  \\
\hline
Midas \cite{lasinger2019towards}  & 78.8 & 80.7 & 75.7 & 63.9 & 57.6 & 83.8 & 87.3 & 86.2 & 54.4 & 78.5 & 68.0 & 83.3 & 82.5 & 78.5 & 78.7 & 48.5 & 77.5 & 71.7 & 79.3 & 73.6 & 74.4   \\
\hline
Boosting \cite{miangoleh2021boosting}  & 77.8 & 80.7 & 76.8 & 64.6 & 60.0 & 81.4 & 86.6 & 85.1 & 54.6 & 76.0 & 67.0 & 84.5 & 82.7 & 78.1 & 78.8 & 47.7 & 76.8 & 75.0 & 80.1 & 73.5 & 74.4   \\
\hline

\end{tabular}
\end{adjustbox}
\end{table*}

\section{Depth Estimation Training Set Experiments}
\label{section:DepEstTrainSetExp}

After choosing the best depth estimator, we investigated the effects of the estimated depth  on Faster R-CNN's performance. While the estimated depth improves Faster R-CNN's performance on SUN RGB-D and Pascal VOC 2007 indoor categories in case of using VGG-16 network, as shown in Tables 4 and 5, it degrades Faster R-CNN's performance on the whole Pascal VOC 2007 dataset, as shown in Table 6. 

This performance loss could be due to the distribution of the training set used for training the depth estimation network; the official pre-trained model was trained with the NYU-D2 dataset which consists of only indoor scenes. Therefore, we trained the depth estimation network  with different combinations of indoor and outdoor datasets to see how training datasets affect the Pascal VOC 2007 object detection results. We used  NYU-D2 for indoor scenes; and Make3D \cite{saxena2006learning, saxena2009make3d} and KITTI for outdoor scenes. In total, we analyzed five different combinations:
\begin{enumerate}
\item NYU-D2 official training split
\begin{itemize}
    \item This includes 249 different scenes and \char`\~45k RGB-D indoor images from NYU-D2 dataset.
\end{itemize}
\item NYU-D2 official training split + Make3D-2 training and test split
\begin{itemize}
    \item This includes 249 different scenes and 498 RGB-D indoor images from NYU-D2 dataset and 374 different scenes and outdoor images from Make3D-2 dataset. To maintain balance of indoor-outdoor data distribution, we chose two random images for each scene in NYU-D2.
\end{itemize}
\item KITTI raw dataset
\begin{itemize}
    \item This includes \char`\~24k RGB-D outdoor images from KITTI raw dataset.
\end{itemize}
\item KITTI raw dataset + NYU-D2 official traning split
\begin{itemize}
\item This includes \char`\~24k RGB-D outdor images from KITTI raw dataset and \char`\~45k indoor images from NYU-D2 dataset.
\begin{enumerate}
\item Mixed Training: we train depth estimation network with mixed indoor and outdoor data.
\item Sequential Training: first, we train depth estimation network with indoor data, then we continue to train it with outdoor data.
\end{enumerate}
\end{itemize}
\end{enumerate}

\begin{figure}
\centering
{
\scriptsize
\renewcommand{\arraystretch}{0.3}
\newcommand{\SKALA}[0]{0.3}
\begin{tabular}{ m{1.5cm}  m{1.2cm} m{1.2cm} m{1.2cm} m{1.2cm}}
\CStart{}Input\CEnd{}         & 
    \includegraphics[scale=\SKALA]{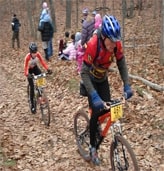} &
    \includegraphics[scale=\SKALA]{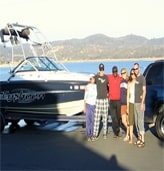} & 
    \includegraphics[scale=\SKALA]{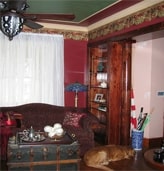} & 
    \includegraphics[scale=\SKALA]{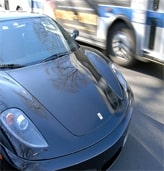} \\\\
\CStart{}Case 1 \CEnd{}         & 
    \includegraphics[scale=\SKALA]{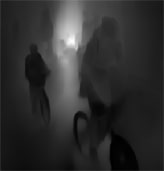} &
    \includegraphics[scale=\SKALA]{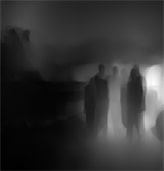} & 
    \includegraphics[scale=\SKALA]{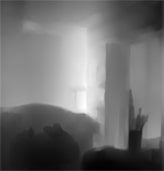} & 
    \includegraphics[scale=\SKALA]{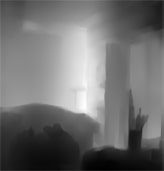} \\\\
\CStart{}Case 2 \CEnd{}         &
    \includegraphics[scale=\SKALA]{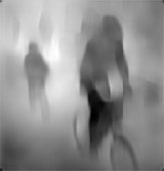} &
    \includegraphics[scale=\SKALA]{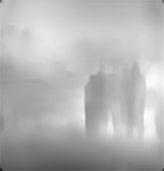} & 
    \includegraphics[scale=\SKALA]{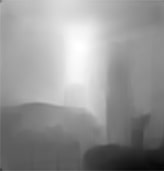} &
    \includegraphics[scale=\SKALA]{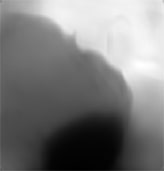} \\\\
\CStart{}Case 3\cite{li2018megadepth} \CEnd{}         & 
    \includegraphics[scale=\SKALA]{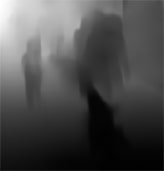} &
    \includegraphics[scale=\SKALA]{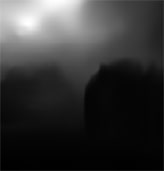} & 
    \includegraphics[scale=\SKALA]{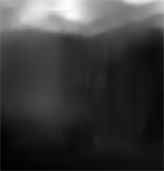} &
    \includegraphics[scale=\SKALA]{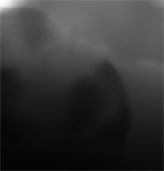} \\\\
\CStart{}Case 4-a\CEnd{}         & 
    \includegraphics[scale=\SKALA]{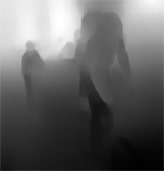} &
    \includegraphics[scale=\SKALA]{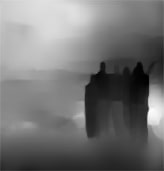} & 
    \includegraphics[scale=\SKALA]{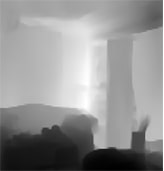} &
    \includegraphics[scale=\SKALA]{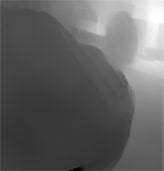} \\\\
\CStart{}Case 4-b\CEnd{}         & 
    \includegraphics[scale=\SKALA]{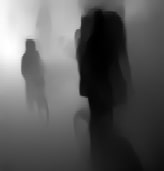} &
    \includegraphics[scale=\SKALA]{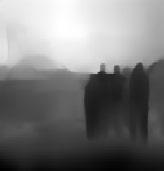} & 
    \includegraphics[scale=\SKALA]{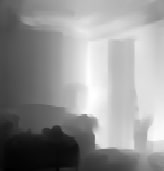} &
    \includegraphics[scale=\SKALA]{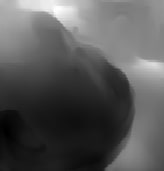} \\\\
\end{tabular}
}
\caption{Depth estimation network trained with different datasets output results. Images are taken from Pascal VOC 2007}
\label{fig:DepTrainEst}
\end{figure}

Visual results for training set combinations are shown in Figure \ref{fig:DepTrainEst} and object detection results are shown in Table 3. Although we observe that changing the training set  generates visually different depth maps, these depth maps degrade the performance of Faster R-CNN on Pascal VOC 2007 (Table 3). This could be related to the outdoor datasets: The Make-3D dataset contains low-quality $86\times107$ depth maps. On the other hand, the KITTI dataset is collected with a LIDAR sensor which generates a very sparse depth map. Creating dense depth maps from such sparse data requires interpolation and this causes incorrect depth information. Because of these outdoor depth data in the training set, estimated depth map, which tends to be very blurry, could not improve outdoor object detection results.

\section{Confusion Matrices}
We analyze confusion matrices of our RGB-D Faster R-CNN and RGB Faster R-CNN to see the effects of using depth. Using ground-truth depth improves all classes' detection performance shown in Table \ref{table:Sun_gt_conf} for SUN RGB-D dataset. These improvements can be divided into two groups: (i) decreasing false negatives, (ii) reducing confusion between classes that have similar RGB appearances such as tv-monitor, table-desk. For the first type of improvement, ground-truth depth helps Faster R-CNN to detect objects which can not be detected using only RGB images. In this way, It decreases false negatives shown in the last column of Table \ref{table:Sun_gt_conf}. Also, it reduces confusion between similar classes in favor of the class with more examples. While desk objects are detected more as tables, table objects are detected less as desks. Television and monitor classes are affected from ground-truth depths in the same way.

Unlike using ground-truth depth, using estimated depth generally increases the number of false negatives. It increases more for Pascal VOC dataset than SUN RGB-D shown in Tables \ref{table:Voc_est_conf} and \ref{table:Sun_est_conf}, because of estimated depth map quality shown in Figure 7.
\begin{table*}[h]
\caption{Confusion matrix difference for ground-truth grayscale depth of SUN RGB-D dataset. Each cell represent the difference between confusion matrix of  (ours) RGB-D Faster R-CNN, and (base model) RGB Faster R-CNN. Bold and red numbers shows improved and failure cases, respectively. FN represent False Negative cases.}
\label{table:Sun_gt_conf}
\centering
\begin{adjustbox}{width=\textwidth}
\small
\begin{tabular}{c|cccccccccccccccccccc}
\hline
  & \rotatebox{90}{bathtub} & \rotatebox{90}{bed} & \rotatebox{90}{b.shelf} & \rotatebox{90}{box} & \rotatebox{90}{chair} & \rotatebox{90}{counter} & \rotatebox{90}{desk} & \rotatebox{90}{door} & \rotatebox{90}{dresser} & \rotatebox{90}{g.bin} & \rotatebox{90}{lamp} & \rotatebox{90}{monitor} & \rotatebox{90}{n.stand} & \rotatebox{90}{pillow} & \rotatebox{90}{sink} & \rotatebox{90}{sofa} & \rotatebox{90}{table} & \rotatebox{90}{t.vision} & \rotatebox{90}{toilet} & \rotatebox{90}{FN} \\
  \hline
bathtub & \textbf{6} & 0 & 0 & 0 & \textbf{-2} & 0 & 0 & \textbf{-1} & 0 & 0 & 0 & 0 & 0 & 0 & \textbf{-2} & 0 & 0 & 0 & 0 & \textbf{-1} \\
\hline
bed & 0 & \textbf{25} & \textbf{-1} & 0 & \textcolor{red}{2} & \textbf{-4} & \textbf{-4} & \textbf{-3} & 0 & \textbf{-1} & 0 & 0 & \textbf{-2} & \textbf{-2} & 0 & \textcolor{red}{2} & \textcolor{red}{3} & \textcolor{red}{1} & 0 & \textbf{-16}\\
\hline
b.shelf & 0 & 0 & \textbf{14} & \textbf{-1} & \textbf{-1} & \textcolor{red}{1} & \textcolor{red}{4} & \textcolor{red}{1} & \textbf{-1} & 0 & 0 & 0 & \textcolor{red}{1} & 0 & 0 & 0 & 0 & \textbf{-1} & 0 & \textbf{-17}\\
\hline
box & 0 & \textbf{-1} & 0 & \textbf{101} & \textbf{-6} & \textbf{-3} & \textcolor{red}{4} & \textbf{-2} & \textbf{-1} & \textbf{-1} & 0 & \textbf{-3} & \textcolor{red}{1} & \textcolor{red}{3} & \textbf{-1} & 0 & \textcolor{red}{4} & \textcolor{red}{2} & 0 & \textbf{-97}\\
\hline
chair & \textbf{-1} & \textbf{-1} & \textbf{-1} & \textcolor{red}{1} & \textbf{459} & \textcolor{red}{1} & \textbf{-14} & \textbf{-1} & 0 & \textbf{-1} & \textcolor{red}{4} & \textbf{-13} & \textcolor{red}{1} & \textbf{-3} & \textbf{-1} & \textcolor{red}{9} & \textbf{-3} & \textcolor{red}{4} & 0 & \textbf{-440}\\
\hline
counter & 0 & 0 & 0 & 0 & \textbf{-1} & \textcolor{red}{
-10} & \textcolor{red}{13} & \textbf{-1} & \textcolor{red}{1} & 0 & 0 & 0 & 0 & 0 & \textbf{-1} & 0 & \textcolor{red}{8} & 0 & 0 & \textbf{-9}\\
\hline
desk & 0 & \textbf{-2} & \textbf{-1} & \textbf{-1} & \textcolor{red}{16} & \textbf{-26} & \textbf{-18} & \textbf{-3} & 0 & \textbf{-2} & \textbf{-4} & 0 & \textcolor{red}{4} & 0 & \textbf{-1} & 0 & \textcolor{red}{126} & \textbf{-3} & 0 & \textbf{-85}\\
\hline
door & \textcolor{red}{1} & 0 & 0 & \textbf{-5} & \textcolor{red}{2} & \textcolor{red}{2} & \textbf{-1} & \textbf{17 }& \textcolor{red}{2} & 0 & \textcolor{red}{1} & 0 & 0 & 0 & 0 & 0 & \textbf{-1} & \textcolor{red}{1} & 0 & \textbf{-19}\\
\hline
dresser & \textbf{-1} & 0 & \textbf{-1} & \textbf{-2} & \textbf{-1} & \textbf{-8} & \textcolor{red}{2} & 0 & \textbf{2} & \textcolor{red}{1} & \textcolor{red}{1} & 0 & \textcolor{red}{16} & 0 & -1 & \textcolor{red}{1} & \textcolor{red}{4} & \textbf{-1} & 0 & \textbf{-12}\\
\hline
g.bin & 0 & \textcolor{red}{1} & 0 & \textcolor{red}{4} & \textbf{-3} & \textbf{-1} & 0 & 0 & \textbf{-1} & \textbf{39} & 0 & 0 & 0 & 0 & \textbf{-1} & \textbf{-1} & \textcolor{red}{2} & \textbf{-1} & 0 & \textbf{-38}\\
\hline
lamp & 0 & 0 & \textcolor{red}{1} & \textcolor{red}{1} & \textcolor{red}{3} & 0 & 0 & \textcolor{red}{1} & \textcolor{red}{2} & \textbf{-1} & \textbf{47} & 0 & 0 & \textcolor{red}{2} & 0 & 0 & \textcolor{red}{2} & 0 & 0 & \textbf{-58}\\
\hline
monitor & 0 & 0 & \textbf{-1} & \textcolor{red}{8} & \textcolor{red}{9} & \textcolor{red}{1} & \textbf{-1} & 0 & 0 & 0 & \textcolor{red}{5} & \textbf{57} & 0 & 0 & 0 & \textcolor{red}{1} & \textcolor{red}{3} & \textbf{-12} & 0 & \textbf{-70}\\
\hline
n.stand & 0 & 0 & 0 & 0 & 0 & \textbf{-1} & 0 & 0 & \textbf{-11} & \textbf{-1} & 0 & 0 & \textbf{18} & \textcolor{red}{-1} & 0 & 0 & \textcolor{red}{8} & \textcolor{red}{1} & 0 & \textbf{-13}\\
\hline
pillow & 0 & \textcolor{red}{1} & 0 & \textcolor{red}{3} & \textbf{-11} & 0 & 0 & 0 & \textbf{-2} & \textbf{-3} & \textbf{-1} & 0 & \textcolor{red}{3} & \textbf{171} & 0 & \textcolor{red}{1} & \textcolor{red}{1} & 0 & 0 & \textbf{-163}\\
\hline
sink & 0 & 0 & 0 & \textcolor{red}{1} & \textbf{-1} & \textbf{-2} & 0 & 0 & 0 & 0 & \textcolor{red}{2} & 0 & 0 & 0 & \textbf{31} & 0 & 0 & 0 & \textbf{-4} & \textbf{-27}\\
\hline
sofa & 0 & \textbf{-2} & 0 & \textbf{-2} & \textbf{-26} & \textcolor{red}{1} & \textbf{-1} & \textcolor{red}{1} & \textbf{-3} & \textcolor{red}{3} & 0 & \textbf{-2} & 0 & \textcolor{red}{1} & 0 & \textbf{58} & \textcolor{red}{2} & \textbf{-1} & 0 & \textbf{-29}\\
\hline
table & \textbf{-1} & \textcolor{red}{3} & 0 & 0 & \textbf{-22} & \textbf{-8} & \textbf{-21} & 0 & \textbf{-1} & 0 & \textbf{-1} & 0 & \textbf{-6} & \textbf{-1} & \textbf{-1} & \textcolor{red}{6} & \textbf{157} & 0 & 0 & \textbf{-104}\\
\hline
t.vision & 0 & 0 & \textcolor{red}{2} & \textcolor{red}{1} & \textcolor{red}{1} & 0 & 0 & \textcolor{red}{1} & \textcolor{red}{2} & 0 & 0 & \textcolor{red}{3} & 0 & 0 & 0 & \textcolor{red}{1} & 0 & \textbf{15} & 0 & \textbf{-26}\\
\hline
toilet & 0 & 0 & 0 & 0 & \textbf{-2} & \textbf{-1} & 0 & 0 & 0 & 0 & 0 & 0 & 0 & 0 & \textcolor{red}{1} & 0 & 0 & 0 & \textbf{3} & \textbf{-1}\\
\hline
\end{tabular}
\end{adjustbox}
\end{table*} 

\begin{table*}[h]
\caption{Confusion matrix difference for estimated grayscale depth of Pascal VOC dataset. Each cell represent the difference between confusion matrix of  (ours) RGB-D Faster R-CNN, and (base model) RGB Faster R-CNN. Bold and red numbers shows improved and failure cases, respectively. FN represent False Negative cases.}
\label{table:Voc_est_conf}
\centering
\begin{adjustbox}{width=\textwidth}
\small
\begin{tabular}{c|ccccccccccccccccccccc}
\hline
  & \rotatebox{90}{A.plane} & \rotatebox{90}{Bicycle} & \rotatebox{90}{Bird} & \rotatebox{90}{Boat} & \rotatebox{90}{Bottle} & \rotatebox{90}{Bus} &  \rotatebox{90}{Car} &  \rotatebox{90}{Cat} &  \rotatebox{90}{Chair} &  \rotatebox{90}{Cow} &  \rotatebox{90}{D.table} &  \rotatebox{90}{Dog} &  \rotatebox{90}{Horse} &  \rotatebox{90}{M.bike} &  \rotatebox{90}{Person} &  \rotatebox{90}{P.plant} &  \rotatebox{90}{Sheep} &  \rotatebox{90}{Sofa} &  \rotatebox{90}{Train} &  \rotatebox{90}{Tv/mon.}&  \rotatebox{90}{FN} \\
 \hline 
A.plane & \textbf{5} & 0 & \textbf{-2} & \textbf{-5} & 0 & 0 & \textbf{-1} & 0 & 0 & 0 & 0 & 0 & 0 & 0 & 0 & 0 & 0 & \textbf{-1} & \textcolor{red}{1} & 0 & \textcolor{red}{3} \\
\hline
Bicycle & 0 & \textbf{12} & 0 & \textcolor{red}{3} & 0 & 0 & \textcolor{red}{1} & 0 & \textbf{-3} & 0 & \textbf{-1} & 0 & 0 & \textbf{-7} & \textbf{-3} & 0 & 0 & 0 & \textbf{-1} & 0 & \textbf{-1} \\
\hline
 Bird & \textcolor{red}{7} & 0 & \textcolor{red}{-1} & \textbf{-1} & 0 & 0 & 0 & \textbf{-3} & 0 & 0 & 0 & \textcolor{red}{4} & \textbf{-4} & 0 & \textcolor{red}{7} & \textbf{-2} & \textbf{-12} & 0 & 0 & 0 & \textcolor{red}{5} \\
\hline
Boat & 0 & \textbf{-1} & \textcolor{red}{1} & \textbf{7} & 0 & \textbf{-2} & \textcolor{red}{1} & 0 & \textbf{-1} & 0 & 0 & 0 & \textbf{-1} & 0 & 0 & 0 & 0 & \textcolor{red}{1} & \textbf{-4} & 0 & \textbf{-1} \\
\hline
Bottle & \textbf{-1} & 0 & \textcolor{red}{2} & 0 & \textcolor{red}{-4} & 0 & \textcolor{red}{1} & 0 & 0 & 0 & 0 & 0 & 0 & 0 & \textbf{-4} & \textbf{-1} & 0 & 0 & 0 & 0 & \textcolor{red}{7} \\
\hline
Bus & 0 & 0 & 0 & 0 & 0 & \textbf{2} & \textbf{-7} & 0 & \textbf{-1} & 0 & 0 & 0 & 0 & \textbf{-1} & \textcolor{red}{1} & 0 & 0 & 0 & 0 & \textcolor{red}{1} & \textcolor{red}{5} \\
\hline
Car & \textbf{-1} & \textcolor{red}{2} & 0 & \textbf{-1} & \textbf{-2} & \textbf{-11} & \textcolor{red}{-14} & 0 & \textcolor{red}{1} & \textbf{-1} & \textbf{-1} & 0 & \textbf{-1} & \textbf{-11} & 0 & 0 & 0 & 1 & \textbf{-3} & \textbf{-3} &  \textcolor{red}{45} \\
\hline
Cat & 0 & 0 & \textbf{-2} & 0 & 0 & 0 & 0 & \textcolor{red}{-5} & \textcolor{red}{7} & 0 & \textbf{-1} & \textcolor{red}{17} & \textbf{-2} & 0 & \textbf{-5} & \textbf{-2} & \textbf{-3 } & \textbf{-3} & 0 & 0 & \textbf{-1} \\
\hline
Chair & 0 & 4 & \textbf{-1} & \textbf{-3} & 0 & 0 & 0 & 0 & \textcolor{red}{-27} & \textcolor{red}{1} & \textbf{-3} & 0 & 0 & \textbf{-1} & \textcolor{red}{10} & \textbf{-2} & 0 & \textbf{-20} & 0 & \textcolor{red}{2} & \textcolor{red}{40} \\
\hline
Cow & 0 & 0 & \textcolor{red}{2} & 0 & 0 & 0 & 0 & 0 & 0 & \textbf{28} & 0 & \textbf{-11} & \textbf{-8} & 0 & \textcolor{red}{1 } & 0 & \textbf{-16} & 0 & 0 & 0 & \textcolor{red}{4} \\
\hline
D.table & \textbf{-1}  & 0 & 0 & \textbf{-1} & 0 & 0 & 0 & 0 & \textbf{-2} & 0 & \textbf{10} & 0 & 0 & 0 & \textbf{-1} & 0 & 0 & \textbf{-12} & 0 & 0 & \textcolor{red}{7} \\
\hline
Dog & 0 & \textcolor{red}{1} & \textcolor{red}{1} & 0 & \textbf{-1} & 0 & 0 & \textbf{-12} & \textcolor{red}{1} & \textcolor{red}{2} & \textbf{-2} & \textbf{9} & 0 & 0 & \textcolor{red}{3} & \textcolor{red}{1} & \textbf{-4} & \textbf{-9} & 0 & 0 & \textcolor{red}{10} \\
\hline
Horse & \textbf{-1} & 0 & \textbf{-1} & 0 & 0 & 0 & 0 & 0 & \textbf{-1} & \textbf{-5} & 0 & \textcolor{red}{1} & \textbf{17} & \textbf{-1} & 0 & 0 & \textbf{-4} & \textbf{-1} & 0 & 0 & \textbf{-4} \\
\hline
M.bike & \textbf{-1} & \textcolor{red}{12} & 0 & 0 & 0 & 0 & \textbf{-12} & 0 & \textbf{-1} & 0 & 0 & 0 & \textbf{-1} & \textbf{2} & \textbf{-2} & 0 & 0 & 0 & \textbf{-3} & 0 & \textcolor{red}{6} \\
\hline
Person & 0 & \textcolor{red}{7} & \textbf{-7} & \textcolor{red}{1} & \textbf{-2} & \textbf{-1} & \textbf{-7} & \textbf{-1} & \textbf{-15} & \textbf{-2} & \textbf{-2} & \textcolor{red}{1} & \textbf{-4} & \textbf{-4} & \textbf{31} & 0 & \textbf{-1} & \textbf{-8} & \textbf{-1} & \textcolor{red}{1} & \textcolor{red}{14} \\
\hline
P.plant & 0 & \textcolor{red}{1} & \textcolor{red}{1} & \textbf{-1} & \textbf{-1} & 0 & 0 & 0 & \textbf{-3} & 0 & \textbf{-1} & 0 & 0 & 0 & 0 & \textcolor{red}{-12} & 0 & \textcolor{red}{1} & 0 & 0 & \textcolor{red}{15} \\
\hline
Sheep & 0 & 0 & \textcolor{red}{3} & 0 & 0 & 0 & \textcolor{red}{1} & \textbf{-1} & \textcolor{red}{1} & \textbf{-4} & 0 & \textbf{-5} & \textcolor{red}{1} & 0 & 0 & 0 & \textbf{6} & \textcolor{red}{1} & 0 & 0 & \textbf{-3} \\
\hline
Sofa & \textcolor{red}{1} & 0 & 0 & 0 & 0 & 0 & \textbf{-3} & 0 & \textbf{-5} & \textcolor{red}{1} & \textbf{-5} & 0 & 0 & 0 & \textcolor{red}{3} & 0 & 0 & \textcolor{red}{-3} & 0 & \textcolor{red}{2} & \textcolor{red}{9} \\
\hline
Train & 0 & 1 & 0 & \textbf{-2} & 0 & 8 & \textbf{-1} & 0 & \textbf{-1} & 0 & 0 & 0 & 0 & \textbf{-1} & 0 & 0 & 0 & 0 & \textcolor{red}{-11} & 0 & \textcolor{red}{7} \\
\hline
TV.{\textbackslash}mon.& 
0 & 0 & 0 & \textbf{-1} & \textcolor{red}{1} & 0 & \textbf{-4} & 0 & \textbf{-9} & 0 & 0 & 0 & 0 & 0 & \textcolor{red}{1} & 0 & 0 & \textcolor{red}{1} & 0 & \textbf{18} & \textbf{-7} \\

\hline
\end{tabular}
\end{adjustbox}
\end{table*}

\begin{table*}[h]
\caption{Confusion matrix difference for estimated grayscale depth of SUN RGB-D dataset. Each cell represent the difference between confusion matrix of  (ours) RGB-D Faster R-CNN, and (base model) RGB Faster R-CNN. Bold and red numbers shows improved and failure cases, respectively. FN represent False Negative cases.}
\label{table:Sun_est_conf}
\centering
\begin{adjustbox}{width=\textwidth}
\small
\begin{tabular}{c|cccccccccccccccccccc}
\hline
  & \rotatebox{90}{bathtub} & \rotatebox{90}{bed} & \rotatebox{90}{b.shelf} & \rotatebox{90}{box} & \rotatebox{90}{chair} & \rotatebox{90}{counter} & \rotatebox{90}{desk} & \rotatebox{90}{door} & \rotatebox{90}{dresser} & \rotatebox{90}{g.bin} & \rotatebox{90}{lamp} & \rotatebox{90}{monitor} & \rotatebox{90}{n.stand} & \rotatebox{90}{pillow} & \rotatebox{90}{sink} & \rotatebox{90}{sofa} & \rotatebox{90}{table} & \rotatebox{90}{t.vision} & \rotatebox{90}{toilet} & \rotatebox{90}{FN} \\
  \hline
bathtub & \textcolor{red}{-3} & 0 & 0 & 0 & \textbf{-2} & 0 & 0 & \textbf{-1} & 0 & \textcolor{red}{1} & 0 & 0 & 0 & 0 & 0 & \textcolor{red}{2} & 0 & 0 & \textcolor{red}{2} & \textcolor{red}{1} \\
\hline
bed & 0 & \textcolor{red}{-1} & \textbf{-1} & 0 & 0 & \textbf{-5} & \textbf{-1} & \textbf{-3} & \textcolor{red}{1} & \textbf{-1} & 0 & 0 & \textbf{-1} & \textbf{-4} & 0 & \textcolor{red}{9} & \textcolor{red}{6} & 0 & 0 & \textcolor{red}{1} \\
\hline
b.shelf & 0 & 0 & \textbf{3} & \textbf{-1} & \textbf{-1} & \textbf{-2} & 1 & \textbf{-5} & \textbf{-1} & 0 & 0 & 0 & 0 & 0 & 0 & 0 & \textbf{-1} & \textbf{-1} & 0 & \textcolor{red}{8} \\
\hline
box & 0 & \textbf{-1} & \textbf{-1} & \textcolor{red}{-18} & \textbf{-7} & \textbf{-4} & \textbf{-3} & \textbf{-3} & \textbf{-1} & \textbf{-5} & 0 & \textbf{-1} & \textbf{-1} & \textbf{-2} & 0 & 0 & \textcolor{red}{9} & \textcolor{red}{1} & 0 & \textcolor{red}{37} \\
\hline
chair & \textbf{-1} & \textbf{-3} & \textbf{-2} & \textbf{-11} & \textcolor{red}{-140} & 0 & \textbf{-7} & \textbf{-2} & \textbf{-2} & \textbf{-1} & 1 & 4 & \textbf{-1} & \textbf{-4} & \textbf{-2} & \textcolor{red}{12} & \textcolor{red}{49} & \textbf{-2} & \textbf{-1} & \textcolor{red}{113} \\
\hline
counter & \textcolor{red}{1} & 0 & \textbf{-2} & 0 & \textbf{-1} & \textcolor{red}{-36} & \textcolor{red}{40} & \textbf{-1} & \textbf{-1} & 0 & 0 & 0 & 0 & 0 & \textbf{-1} & 0 & \textcolor{red}{19} & 0 & 0 & \textbf{-18} \\
\hline
desk & 0 & \textbf{-2} & \textbf{-7} & \textbf{-8 }& \textbf{-39} & \textbf{-44} & \textbf{99} & \textbf{-4} & \textbf{-3} & 0 & \textbf{-2} & 0 & 0 & 0 & \textbf{-1} & \textbf{-1} & \textcolor{red}{279} & \textbf{-3} & 0 & \textbf{-264} \\
\hline
door & \textcolor{red}{1} & 0 & 0 & \textbf{-5} & 0 & 0 & 0 & \textcolor{red}{-43} & \textcolor{red}{2} & 0 & 0 & \textcolor{red}{1} & 0 & 0 & 0 & 0 & \textbf{-2} & 0 & 0 & \textcolor{red}{46} \\
\hline
dresser & \textbf{-1} & \textcolor{red}{1} & \textbf{-1} & \textbf{-1} & \textbf{-1} & \textbf{-7} & \textcolor{red}{1} & 0 & \textcolor{red}{-15} & \textcolor{red}{2} & \textcolor{red}{1} & 0 & \textcolor{red}{8} & 0 & 0 & 0 & \textcolor{red}{6} & 0 & 0 & \textcolor{red}{7} \\
\hline
g.bin & 0 & \textbf{-2} & 0 & \textbf{-2 }& \textbf{-6 }& \textbf{-2} & 0 & \textbf{-1} & 0 & \textcolor{red}{-17 }& 0 & 0 & \textbf{-1} & \textcolor{red}{1} & \textbf{-1} & 0 & \textcolor{red}{2} & \textbf{-1} & 0 & \textcolor{red}{30} \\
\hline
lamp & 0 & \textbf{-1} & 0 & \textbf{-2} & 0 & 0 & 0 & \textbf{-2} & 0 & \textbf{-1} & \textcolor{red}{-7 }& \textbf{-1 }& 0 & \textcolor{red}{1} & 0 & \textcolor{red}{1} & 0 & \textbf{-1} & 0 & \textcolor{red}{13} \\
\hline
monitor & 0 & 0 & \textbf{-1} & \textbf{-3} & \textbf{-1} & 0 & \textbf{-1} & 0 & 0 & \textbf{-1} & 0 & \textbf{26} & 0 & 0 & 0 & 0 & 0 & \textcolor{red}{3} & 0 & \textbf{-22} \\
\hline
n.stand & 0 & 0 & 0 & \textbf{-4} & \textbf{-1} & \textbf{-2} & \textcolor{red}{3} & 0 & \textbf{-6} & \textbf{-1} & 0 & 0 & \textbf{4} & 0 & 0 & 0 & \textcolor{red}{5} & \textbf{-1} & 0 & \textcolor{red}{3} \\
\hline
pillow & 0 & \textcolor{red}{2} & 0 & \textbf{-2} & \textbf{-10 }& 0 & 0 & 0 & \textbf{-2} & \textbf{-4} & 0 & 0 & 0 & \textcolor{red}{-3} & 0 & 0 & \textcolor{red}{1} & 0 & 0 & \textcolor{red}{18} \\
\hline
sink & \textbf{-1} & 0 & 0 & 0 & \textbf{-1} & \textbf{-3} & 0 & 0 & 0 & \textcolor{red}{2} & \textcolor{red}{1} & 0 & 0 & 0 & \textbf{1} & 0 & \textcolor{red}{3} & 0 & \textbf{-2} & 0 \\
\hline
sofa & 0 & \textbf{-9 }& 0 & \textbf{-4 }& \textbf{-32} & \textcolor{red}{2} & \textbf{-1} & 0 & \textbf{-3} & 0 & 0 & 0 & 0 & \textcolor{red}{1} & 0 & \textbf{72} & \textbf{-4 }& \textbf{-2} & 0 & \textbf{-20} \\
\hline
table & \textcolor{red}{1} & \textcolor{red}{1} & \textbf{-1} & \textbf{-9} & \textbf{-53} & \textbf{-23} & \textcolor{red}{82} & 0 & \textbf{-1} & \textcolor{red}{1} & \textcolor{red}{1} & 0 & \textbf{-4} & \textbf{-1 }& 0 & \textcolor{red}{2} & \textbf{528} & 0 & \textbf{-1} & \textbf{-523} \\
\hline
t.vision & 0 & 0 & 0 & \textbf{-2} & 0 & 0 & \textcolor{red}{1} & \textcolor{red}{2} & \textcolor{red}{1} & 0 & 0 & \textcolor{red}{8} & 0 & 0 & 0 & 0 & 0 & \textbf{-4} & 0 & \textbf{-6}\\
\hline
toilet & 0 & 0 & 0 & 0 & 0 & \textbf{-1} & 0 & 0 & 0 & \textcolor{red}{1} & \textcolor{red}{1} & 0 & 0 & 0 & \textcolor{red}{1} & 0 & 0 & 0 & \textcolor{red}{-2} & 0 \\
\hline
\end{tabular}
\end{adjustbox}
\end{table*} 
\clearpage

\bibliography{supplement}